\PassOptionsToPackage{unicode}{hyperref}
\PassOptionsToPackage{hyphens}{url}
\PassOptionsToPackage{dvipsnames,svgnames,x11names}{xcolor}
\documentclass[
  12pt]{article}

\usepackage{amsmath,amssymb}
\usepackage{iftex}
\usepackage{mathtools} 
\allowdisplaybreaks     

\ifPDFTeX
  \usepackage[T1]{fontenc}
  \usepackage[utf8]{inputenc}
  \usepackage{textcomp} 
\else 
  \usepackage{unicode-math}
  \defaultfontfeatures{Scale=MatchLowercase}
  \defaultfontfeatures[\rmfamily]{Ligatures=TeX,Scale=1}
\fi
\usepackage{lmodern}
\ifPDFTeX\else  
\fi
\IfFileExists{upquote.sty}{\usepackage{upquote}}{}
\IfFileExists{microtype.sty}{
  \usepackage[]{microtype}
  \UseMicrotypeSet[protrusion]{basicmath} 
}{}
\makeatletter
\@ifundefined{KOMAClassName}{
  \IfFileExists{parskip.sty}{%
    \usepackage{parskip}
  }{
    \setlength{\parindent}{0pt}
    \setlength{\parskip}{6pt plus 2pt minus 1pt}}
}{
  \KOMAoptions{parskip=half}}
\makeatother
\usepackage{xcolor}
\makeatletter
\ifx\paragraph\undefined\else
  \let\oldparagraph\paragraph
  \renewcommand{\paragraph}{
    \@ifstar
      \xxxParagraphStar
      \xxxParagraphNoStar
  }
  \newcommand{\xxxParagraphStar}[1]{\oldparagraph*{#1}\mbox{}}
  \newcommand{\xxxParagraphNoStar}[1]{\oldparagraph{#1}\mbox{}}
\fi
\ifx\subparagraph\undefined\else
  \let\oldsubparagraph\subparagraph
  \renewcommand{\subparagraph}{
    \@ifstar
      \xxxSubParagraphStar
      \xxxSubParagraphNoStar
  }
  \newcommand{\xxxSubParagraphStar}[1]{\oldsubparagraph*{#1}\mbox{}}
  \newcommand{\xxxSubParagraphNoStar}[1]{\oldsubparagraph{#1}\mbox{}}
\fi
\makeatother

\usepackage{longtable,booktabs,array}
\usepackage{calc} 
\usepackage{etoolbox}
\makeatletter
\patchcmd\longtable{\par}{\if@noskipsec\mbox{}\fi\par}{}{}
\makeatother
\IfFileExists{footnotehyper.sty}{\usepackage{footnotehyper}}{\usepackage{footnote}}
\makesavenoteenv{longtable}
\usepackage{graphicx}
\makeatletter
\def\maxwidth{\ifdim\Gin@nat@width>\linewidth\linewidth\else\Gin@nat@width\fi}
\def\maxheight{\ifdim\Gin@nat@height>\textheight\textheight\else\Gin@nat@height\fi}
\makeatother
\setkeys{Gin}{width=\maxwidth,height=\maxheight,keepaspectratio}
\makeatletter
\def\fps@figure{htbp}
\makeatother

\makeatletter
\@ifpackageloaded{caption}{}{\usepackage{caption}}
\AtBeginDocument{%
\ifdefined\contentsname
  \renewcommand*\contentsname{Table of contents}
\else
  \newcommand\contentsname{Table of contents}
\fi
\ifdefined\listfigurename
  \renewcommand*\listfigurename{List of Figures}
\else
  \newcommand\listfigurename{List of Figures}
\fi
\ifdefined\listtablename
  \renewcommand*\listtablename{List of Tables}
\else
  \newcommand\listtablename{List of Tables}
\fi
\ifdefined\figurename
  \renewcommand*\figurename{Figure}
\else
  \newcommand\figurename{Figure}
\fi
\ifdefined\tablename
  \renewcommand*\tablename{Table}
\else
  \newcommand\tablename{Table}
\fi
}
\@ifpackageloaded{float}{}{\usepackage{float}}
\floatstyle{ruled}
\@ifundefined{c@chapter}{\newfloat{codelisting}{h}{lop}}{\newfloat{codelisting}{h}{lop}[chapter]}
\floatname{codelisting}{Listing}

\makeatother
\makeatletter
\@ifpackageloaded{caption}{}{\usepackage{caption}}
\@ifpackageloaded{subcaption}{}{\usepackage{subcaption}}
\makeatother

\ifLuaTeX
  \usepackage{selnolig}  
\fi
\usepackage[]{natbib}
\usepackage{bookmark}

\IfFileExists{xurl.sty}{\usepackage{xurl}}{} 
\hypersetup{
  pdftitle={Title},
  pdfauthor={Author 1; Author 2},
  pdfkeywords={3 to 6 keywords, that do not appear in the title},
  colorlinks=true,
  linkcolor={blue},
  filecolor={Maroon},
  citecolor={Blue},
  urlcolor={Blue},
  pdfcreator={LaTeX via pandoc}}

\newcommand{\anon}{1}

\usepackage{amsmath}
\usepackage{graphicx,multirow}
\usepackage{subcaption}
\usepackage{enumerate}
\usepackage{natbib}
\usepackage{url} 
\usepackage{xcolor}
\usepackage{array}
\newcolumntype{B}{>{\color{blue}}c}

\usepackage{enumitem}
\usepackage[utf8]{inputenc}
\usepackage{bm}
\usepackage{bbm,breqn}
\usepackage{amsmath}
\usepackage{float,color}
\usepackage{amssymb,amsthm}

\newtheorem{theorem}{Theorem}

\newtheorem{definition}{Definition}

\newtheorem{proposition}{Proposition}

\newtheorem{assumption}{Assumption}

\newcommand{\dN}{\mathcal{N}}

\newcommand{\E}{\mathrm{E}}
\newcommand{\Var}{\mathrm{Var}}
\newcommand{\Cov}{\mathrm{Cov}}

\begin{document}

\def\spacingset#1{\renewcommand{\baselinestretch}%
{#1}\small\normalsize} \spacingset{1}


\date{} 

\if1\anon
{
  \title{\bf Gaussian Invariant  Markov Chain Monte Carlo}
 \author{Michalis K. Titsias
    \hspace{.2cm}\\
    Google DeepMind, UK
    and \\
    Angelos Alexopoulos \\
    Athens University of Economics and Business, Greece
    and \\
    Siran Liu \\
    UCL, UK
     and \\
    Petros Dellaportas \\
    UCL, UK and Athens University of Economics and Business, Greece}
  \maketitle
} \fi

\if0\anon
{
  \bigskip
  \bigskip
  \bigskip
  \begin{center}
    {\LARGE\bf Gaussian Invariant  Markov Chain  Monte Carlo}
\end{center}
  \medskip
} \fi

\bigskip
\begin{abstract}
We develop sampling methods,  which consist of  Gaussian invariant versions of random walk Metropolis (RWM), Metropolis adjusted Langevin algorithm (MALA) and second order Hessian or Manifold MALA. 
Unlike standard RWM and MALA, we show that Gaussian invariant sampling can lead to ergodic estimators with 
improved statistical efficiency. This is due to a remarkable property of
Gaussian invariance that allows us to obtain exact analytical solutions to the Poisson equation
for Gaussian targets. These solutions can be used to construct efficient  and easy to use control variates for variance reduction
of estimators under any intractable target. We demonstrate the new samplers and estimators in several examples, 
including high dimensional targets in latent Gaussian  models where we compare against several advanced methods and obtain 
state-of-the-art results. 
We also provide theoretical results regarding geometric ergodicity, and an optimal scaling analysis that shows the dependence of the optimal acceptance rate on the Gaussianity of the target. 
\end{abstract}

\noindent%
\vfill

\spacingset{1.2} 

 
\section{Introduction}

In Bayesian statistics a popular approach to sampling-based inference relies on Markov chain Monte Carlo (MCMC) which constructs an ergodic Markov chain $\{X_0,X_1,\ldots,X_{n-1}\}_{n \geq 0}$ with continuous state space $\mathcal{X} \subseteq \Re^d$, transition kernel $P$ and invariant measure $\pi$, referred to as the \emph{target}, for which we assume there is a density denoted by $\pi(x)$. The expected value $\E_{\pi}[F] := \pi(F) = \int F(x) \pi(x) dx$ of a real-valued function $F$ 
is estimated through the ergodic average
\begin{equation}
\label{eq:std_estimator}
\mu_n(F) = \frac{1}{n}\sum_{i=0}^{n-1}F(X_i),\,\,\, X_i \in \mathcal{X}
\end{equation}
which satisfies, for any initial distribution of $X_0$, a central limit theorem of the form
\begin{equation}
\label{eq:CLT}
\sqrt{n}\big(\mu_n(F) - \pi(F)\big)  
\overset{D}{\to} \mathcal{N}\Big(0,\sigma^2_F\Big), 
\end{equation}
with the asymptotic variance $\sigma^2_F$ given by 
\begin{equation}
\label{eq:s2F}
\sigma^2_F := \Var_\pi[F(X_0)] + 2\sum_{i=0}^{\infty}\Cov_\pi\{F(X_{0}),F(X_{i+1})\}.
\end{equation}
Different MCMC algorithms are usually compared with respect to their computational complexity and their statistical efficiency which is measured by the size of the asymptotic variance $\sigma^2_F$.  We focus on  the popular Metropolis-Hastings (MH) algorithm, where at each iteration, given the current state $x$, a candidate state $y$ is drawn from a proposal distribution  $q(y|x)$ and it is accepted or rejected with probability 
\begin{equation*}
\label{eq:alpha}
\alpha(x,y) =  \min \left\{ 1, \frac{\pi(y)q(x|y)}{\pi(x)q(y|x)} \right\}.
\end{equation*}
This  results in a transition kernel with density
\begin{equation}
P(y| x) = \alpha(x,y)q(y|x) + \left( 1 - \int
\alpha(x,z)q(z|x)dz \right) \delta_x(y),
\label{eq:transition_MH}
\end{equation}
where $\delta_x$ is a point mass centred at $x$. 
Two widely-used samplers are the random walk Metropolis (RWM) with  proposal 
\begin{equation}
q(y|x)  = \mathcal{N}(y|x, 2 \gamma \Sigma ),
\label{eq:RWM}
\end{equation}
and 
the Metropolis adjusted Langevin algorithm 
(MALA) with proposal 
\begin{equation}
q(y|x)  = \mathcal{N}(y|x + \gamma 
\Sigma \nabla \log \pi(x), 2 \gamma \Sigma),
\label{eq:MALA}
\end{equation}
where  $\mathcal{N}(y|\mu, \Sigma )$ denotes the multivariate Gaussian density with mean $\mu$ and covariance matrix $\Sigma$.  In (\ref{eq:RWM}) and (\ref{eq:MALA}) we write the covariance matrix as $2 \gamma \Sigma$ to 
make it comparable with the notation in Section \ref{sec:Gauss_inv_mcmc} where $\gamma > 0$ is a global step size 
and $\Sigma$ is a preconditioning covariance matrix. There exist also generalizations of MALA 
where the constant preconditioner 
$\Sigma$ is replaced by a state-dependent 
matrix 
$A_x$ such as an inverse Fisher matrix or the negative inverse Hessian matrix \citep{girolami2011riemann, petra2014computational}.

The main motivation of our work is that ergodic estimation based on  RWM and MALA %
can be sub-optimal due to lack of invariance to Gaussian distributions. 
Specifically,  
these algorithms do not have the Gaussian invariance property defined as follows: 
\begin{definition} (Gaussian invariance) Given a Gaussian target $\pi(x) = \mathcal{N}(x|\mu, \Sigma)$ 
an MH sampler is Gaussian invariant if the proposal 
distribution is reversible to $\mathcal{N}(x|\mu, \Sigma)$, i.e., it holds $\mathcal{N}(y|\mu, \Sigma) q(x|y) =  
\mathcal{N}(x|\mu, \Sigma) q(y|x)$.
In such a case the acceptance probability is always $\alpha(x,y)=1$, so that  all proposed samples are accepted.
\end{definition}
The absence of Gaussian invariance 
of RWM and MALA  
can result in reduced statistical efficiency for two reasons: 

\begin{enumerate}
\item[(a)] When the target $\pi$ becomes close to a Gaussian distribution these algorithms do not recover i.i.d.\ sampling. For instance, in the limiting case when the target is  
$\mathcal{N}(x|\mu, \Sigma)$ there is no value of $\gamma > 0$,  in either RWM or MALA, that results in i.i.d.\ sampling from $\mathcal{N}(x|\mu, \Sigma)$. 
MALA comes closer to achieving i.i.d.\ sampling since by setting $\gamma=1$ the proposal in \eqref{eq:MALA} becomes the independent Metropolis proposal 
 $q(y) = \mathcal{N}(y|\mu, 2 \Sigma)$ which has the same mean $\mu$ as the target, but still twice the target covariance.

\item[(b)] For a Gaussian target $\mathcal{N}(x|\mu, \Sigma)$ a Gaussian invariant kernel can achieve zero variance estimation, i.e., it can minimize the asymptotic variance in \eqref{eq:s2F},  for many functions $F(x)$ for which 
the integral $\int F(x) \mathcal{N}(x|\mu, \Sigma) d x$ is analytic.  
This is achieved by solving  
the Poisson equation, see e.g., \cite{Meyn2009MarkovEdition}, and obtain a control variate that results in a zero variance ergodic estimator for $E_\pi[F]$.
In contrast, when applying RWM or MALA  to sample from  $\mathcal{N}(x|\mu, \Sigma)$, it seems infeasible to obtain 
zero variance estimation since as shown by \cite{alexopoulos2023variance} 
we cannot solve analytically the Poisson equation, not even 
for the simplest function $F(x)=x$. 
This implies that efficient variance reduction for non-Gaussian 
targets when running RWM or MALA can be very hard, since it is already difficult for the tractable Gaussian target case. 
\end{enumerate}

Motivated by the above, we develop Gaussian invariant (GI)  samplers that are direct modifications of RWM, MALA and second order Hessian or manifold MALA \citep{girolami2011riemann,petra2014computational}. 
We refer to the new  samplers
as GI-RWM and GI-MALA. 
As a main application of these samplers, we consider latent Gaussian models which are widely used in various areas of statistics, 
machine learning and Bayesian inverse problems 
\citep{Rueetal2009,rasmussen2006gaussian,cotter2013mcmc}. 
In such applications, the simplest  Gaussian invariant sampler is the preconditioned Crank Nicolson (pCN) 
\citep{cotter2013mcmc, beskos2008mcmc,Neal99},  
while the most advanced samplers, that we introduce, construct second order Hessian approximations that can outperform state-of-the-art sampling in latent Gaussian models \citep{titsias2018auxiliary}.

In a second contribution 
we follow previous 
work \citep{dellaportas2012control,alexopoulos2023variance,douc2022solving,liu2024can}, and construct 
control variates  through approximate solutions of the Poisson equation 
in order to reduce the variance of Monte Carlo estimators. More precisely, for Gaussian invariant samplers we  derive  exact solutions to the Poisson equation
for several functions $F$ and Gaussian targets. Then, we use these solutions  
to increase the sampling efficiency of the ergodic estimators under intractable non-Gaussian targets, by applying variance reduction using control variates.

Furthermore, we establish theoretical results demonstrating the geometric ergodicity of our samplers
by following 
\cite{roberts1996exponential,roberts1996geometric} and \cite{roy2023convergence}.
When an analytical solution to the Poisson equation is unavailable,  we produce an approximate solution and prove its convergence properties. Finally, we study in detail the optimal scaling of GI-MALA.  While previous studies on the optimal scaling of MCMC algorithms 
assume that the dimension of the target 
tends to infinity \citep{gelman1997weak,roberts1998optimal,roberts2001optimal},  such an assumption yields a trivial solution in our setting. Therefore, we obtain an approximate optimal scaling under a finite dimension and characterize its dependence on deviations from the Gaussianity of the target.

\subsection{Outline}
\label{sec:outline}

The paper is structured as follows. In Section 
\ref{sec:Gauss_inv_mcmc}
we present 
the general form of  Gaussian invariant samplers
that incorporate different levels of information from the target such as first and second order derivatives. 
In Section \ref{sec:reduced_variance_estimation} we develop a variance reduction estimation framework 
that utilizes Gaussian invariant samplers and control variates. 
In Section \ref{sec:MCMC_GPs}
we specialize our framework 
to the latent Gaussian models and in Section \ref{sec:experiments} we provide experimental results.
In Section \ref{sec:theory} we describe  
theoretical results, 
such as a finite dimension optimal scaling analysis, 
and finally we conclude the paper with a discussion in Section \ref{sec:discussion}.

\section{Gaussian invariant MCMC}
\label{sec:Gauss_inv_mcmc}

In this section we develop Gaussian invariant MCMC proposal distributions based on three levels of  increasing computational cost: (i) a zero-order RWM-like 
proposal, (ii) a first-order or MALA-like  proposal 
that exploits gradients
$\nabla \log \pi(x)$ of the 
log target, and (iii) 
a second-order proposal that 
automatically specifies the preconditioning by possibly exploiting some information of the Hessian matrix $\nabla^2 \log \pi(x)$.  

To start with  case (i), suppose for $\pi(x)$ 
we have a Gaussian approximation $\hat{\pi}(x) = \mathcal{N}(x|\mu, \Sigma)$. A way to use $\hat{\pi}$ to target $\pi$ is to 
consider the independent Metropolis algorithm with proposal $q(y) = \hat{\pi}(y)$. However, this can have high rejection rates when $\hat{\pi}$ is not a good approximation to $\pi$. Another way is to use a RWM-like scheme with the following 
autoregressive proposal,  
\begin{equation}
    \label{eq:proposal_1}
   q(y|x) = \dN(y|(1-\gamma) x + \gamma \mu, (2\gamma - \gamma^2) \Sigma),
\end{equation}
where the parameter $\gamma \in (0, 2)$ controls the autocorrelation between $y$ and $x$ 
so that when $\gamma=1$ the proposal 
reduces to independent Metropolis, while when $\gamma$ is close to zero,  $y$ will be  close to $x$. 
 It is easy to verify that the above $q(y|x)$ 
is reversible and therefore invariant to the Gaussian approximation $\hat{\pi}(x) = \mathcal{N}(x|\mu, \Sigma)$.
Also, it has the same form as pCN that has been applied to latent Gaussian models 
\citep{cotter2013mcmc,Neal99, beskos2008mcmc}. The difference  is that here  we care about 
invariance over a Gaussian approximation  $\hat{\pi}$ to some generic target $\pi$, where $\pi$ is not necessarily  a latent Gaussian  model.  

A limitation of  \eqref{eq:proposal_1} 
is that it depends on specific values for both parameters 
$\mu$ and $\Sigma$. We can 
avoid this  by adding information from the target $\pi$, such as first and second order derivatives. In case (ii) we use the gradient vector $\nabla \log \pi(x)$ and construct the proposal
\begin{equation}
    \label{eq:proposal_2}
q(y|x) = \dN(y|x + \gamma \Sigma \nabla \log \pi(x), (2\gamma - \gamma^2) \Sigma),
\end{equation}
which depends only on  $\Sigma$ and not on $\mu$. 
Now observe that for Gaussian target $\mathcal{N}(x|\mu,\Sigma)$, where $\mu$ is arbitrary but $\Sigma$ is the same as in \eqref{eq:proposal_2}, the mean of the above proposal simplifies as 
$x+ \gamma\Sigma \nabla \log \mathcal{N}(x|\mu,\Sigma) = 
x + \gamma \Sigma \Sigma^{-1} (\mu - x) = (1-\gamma) x + \gamma \mu$,
so the above proposal 
reduces to the Gaussian invariant proposal from \eqref{eq:proposal_1}. 
However, unlike \eqref{eq:proposal_1},   \eqref{eq:proposal_2} enjoys the advantage that does not require specifying  $\mu$, and therefore it is reversible  over the full family of Gaussians having a certain 
covariance matrix $\Sigma$, but arbitrary mean $\mu$.

In case (iii) we want to additionally avoid having to
 specify  $\Sigma$, and for that we need more information from the target $\pi$. To formalize this we define  the following preconditioning.
 \begin{definition}
   The matrix $A_x := A(x, \pi)$, that is a function of $x$ and $\pi$, is called a Gaussian optimal preconditioning if it is a symmetric positive definite  matrix and for Gaussian targets $\pi(x)=\mathcal{N}(x|\mu, \Sigma)$ it holds 
 $A(x, \mathcal{N}(x|\mu, \Sigma)) = \Sigma$.
 \label{def:preconditioning}
 \end{definition}
Using $A_x$ we can consider the  
 proposal distribution 
\begin{equation}
    \label{eq:proposal_3}
q(y|x) = \dN(y|x + \gamma A_x \nabla \log \pi(x), (2 \gamma - \gamma^2) A_x),
\end{equation}
which is the same as \eqref{eq:proposal_2} but  does not depend on $\Sigma$; instead it uses $A_x$ which is a function of current state $x$ and $\pi$. 
Due to the Gaussian optimality of $A_x$,  this proposal is invariant to any 
Gaussian  
$\dN(x|\mu, \Sigma)$, 
since then $A_x=\Sigma$ and the proposal reduces to the one from \eqref{eq:proposal_2}.  

 One way to specify $A_x$ is to make it equal to 
 the negative of the inverse  Hessian of the log target, i.e., $A_x^{-1} = -\nabla^2 \log\pi(x)$. This has been popularized by the stochastic Newton MCMC algorithm of
\cite{petra2014computational}, 
although this idea appears  earlier  
\citep{qi2002hessian-based}. The MCMC proposal in  \cite{petra2014computational} is a special case of \eqref{eq:proposal_3} obtained for $\gamma=1$. The manifold MALA algorithm in \cite{girolami2011riemann}
also uses a state-dependent preconditioning that is often close or identical to the negative inverse Hessian. However, this manifold  MALA is never a special case of \eqref{eq:proposal_3} as any MALA is not Gaussian invariant.
.

When the Hessian is expensive to compute, we  
can use some approximation to the Hessian as long as 
 for Gaussian targets
 it returns the exact $\Sigma$. For instance, in Section \ref{sec:MCMC_GPs} where we specialize the proposal \eqref{eq:proposal_3} to latent Gaussian models,
 we use computationally 
efficient approximations to the Hessian which result in state-of-the-art 
performance outperforming many  previous samplers. 
Also, the preconditioning $A_x$  does not need to be state-dependent 
to be compatible with  Definition \ref{def:preconditioning}. For example, the constant preconditioning for MALA in \cite{titsias2023optimal},  having the inverse Fisher form
$A = \mathcal{I}^{-1}$ with  $\mathcal{I} = \E_{\pi(x)} [ \nabla \log \pi (x)  \nabla \log \pi(x)^\top]$, is a suitable option since for a Gaussian 
target the  Fisher 
is $\mathcal{I} = \Sigma^{-1}$. The covariance matrix of the target $\pi$ is also a valid option. This suggests that in practice  we could use adaptive MCMC algorithms \citep{haario2001adaptive, titsias2023optimal} to 
estimate a constant $A$. 
 
By comparing the proposals \eqref{eq:proposal_1} 
and \eqref{eq:proposal_2}-\eqref{eq:proposal_3} with the expressions for RWM and MALA from \eqref{eq:RWM}-\eqref{eq:MALA} 
we observe that the connection especially with MALA
is really striking.  The MALA proposal
in \eqref{eq:MALA} 
and the Gaussian invariant 
gradient-based proposal 
in \eqref{eq:proposal_2}
have the same mean  
$x + \gamma \Sigma \nabla \log \pi(x)$ and differ 
only in the multiplicative scalar in front of $\Sigma$: MALA uses $2\gamma$ while \eqref{eq:proposal_2}
uses a smaller scalar equal to 
$2 \gamma - \gamma^2$.    
Despite these differences, 
we shall refer to 
\eqref{eq:proposal_1} 
as Gaussian invariant RWM (GI-RWM)
and similarly to \eqref{eq:proposal_2}-\eqref{eq:proposal_3} 
as GI-MALA. 

\section{Reduced variance estimation
\label{sec:reduced_variance_estimation}
}

Here, we introduce variance reduction estimators for Gaussian invariant MCMC using control variates and the Poisson equation  \citep{andradottirelal93,Meyn2009MarkovEdition,dellaportas2012control}. In Section \ref{sec:control_variates_poisson}, we give a brief introduction to the Poisson equation. In Section \ref{sec:exact_solutions},  
we obtain exact solutions to the Poisson equation for Gaussian targets and Gaussian invariant proposals. In Section \ref{sec:variance_reduction}, 
we use the former 
solutions as control variates for  variance reduction in intractable non-Gaussian targets. In Section 
\ref{sec:algo_summary},
we provide an algorithmic  
summary and guidelines 
for implementing  the estimators.

\subsection{Control variates and the Poisson equation
\label{sec:control_variates_poisson}
} 

Throughout this section we assume a general, i.e., possibly non-Gaussian, target $\pi$ and any  $\pi$-invariant transition kernel $P$.

A natural control variate in ergodic estimators of Markov chains is a function of the form 
$$
U(x) = \int P(y|x) G(y) d y - G(x)
= PG(x) - G(x)
$$ 
where $PG(x) = \int P(y|x) G(y) d y = \E [ G(X_1)| X_0 = x]$. 
Since  the transition kernel density $P(y|x)$ is $\pi$-invariant, i.e., 
$\pi(y) = \int P(y|x) \pi(x) d x$, 
we have   $\E_\pi[U(x)] = 0$. Then for samples  
$X_i$ drawn from $\pi$ we can modify
the estimator in \eqref{eq:std_estimator}
by adding the control variates  $U(X_i)$  
and construct the unbiased estimator
\begin{equation}
\label{eq:modified_estimator}
\mu_{n,G}(F) = \frac{1}{n}\sum_{i=0}^{n-1}\{F(X_i) + P G(X_i) - G(X_i)\}.  
\end{equation}
For the MH kernel in \eqref{eq:transition_MH} $PG$ is written as 
\begin{align}
PG(x) & = \int \alpha(x, y)q(y|x)G(y)dy + \left(1- \int \alpha(x,z)q(z|x)dz \right) G(x) \nonumber \\
    & = G(x) + \int \alpha(x,y)(G(y) - G(x))q(y|x)dy \label{PG}
\end{align}
and the estimator simplifies  
as
\begin{equation}
\label{eq:modified_estimatorMH}
\mu_{n,G}(F) = \frac{1}{n}\sum_{i=0}^{n-1}\left\{F(X_i) + \int \alpha(X_i,y)(G(y) - G(X_i))q(y|X_i)dy\right\}.
\end{equation}

To choose a suitable function $G$ in the estimator \eqref{eq:modified_estimator}
we consider the Poisson equation, that characterizes the optimal choice resulting in zero variance. 
This equation says that there is a function 
$\hat{F}(x)$ that satisfies 
\begin{equation}
\int P(y|x) \hat{F}(y) d y - \hat{F}(x) = - F(x) + \pi(F),  \ \ \forall  x \sim \pi(x)
\label{eq:poisson}
\end{equation}
or with shorthand notation 
$
P \hat{F}(x) - \hat{F}(x) = - F(x) + \pi(F). 
$
We observe that if $\hat{F}$ is a solution then 
$\hat{F} + c$ is also a solution, thus in practice  additive constants can be ignored.  If we know $\hat{F}$ 
 we can use $G= \hat{F}$ 
in \eqref{eq:modified_estimator} and obtain 
\begin{equation}
\label{eq:modified_estimator_optimal}
\mu_{n,G}(F) = \frac{1}{n}\sum_{i=0}^{n-1}\{F(X_i) + P \hat{F}(X_i) - \hat{F}(X_i) \}
= \frac{1}{n}\sum_{i=0}^{n-1}\{F(X_i) 
- F(X_i) + \pi(F) 
\}
\} = \pi(F),
\end{equation}
so this estimator is exact, i.e., it has zero variance. 
A theoretical way to write the solution is as the following infinite series
\citep{dellaportas2012control},
\begin{equation}
\hat{F}(x) = F(x) - \pi(F) + \sum_{n=1}^{\infty} \left( P^n F(x) - \pi(F) \right),
\label{eq:poisson_inf_series}
\end{equation}
where $P^n F(x) = \E[F(X_n) | X_0=x]
\xrightarrow{n \rightarrow \infty} \pi(F)$. 
Unfortunately,  
this expression does not provide an analytical closed-form approximation (e.g., obtained by truncating the infinite series)
since for general targets and MH kernels  the $n$-step expectations  $P^n F(x)$ are intractable. 
However, as shown next in Section \ref{sec:exact_solutions} for Gaussian invariant kernels, Gaussian targets and several functions $F(x)$, 
we can obtain analytical solutions. In Section \ref{sec:variance_reduction} we will use these solutions to construct control variates for non-Gaussian targets. 

\subsection{Exact solutions for Gaussian targets
\label{sec:exact_solutions}
}

Throughout this section we assume 
a Gaussian target $\pi(x) = \mathcal{N}(x|\mu,\Sigma)$ and a Gaussian invariant kernel based on either GI-RWM  from \eqref{eq:proposal_1} or GI-MALA from 
\eqref{eq:proposal_2}-\eqref{eq:proposal_3}.

\begin{proposition} 
Suppose for the target 
$\mathcal{N}(x|\mu, \Sigma)$ we apply MH using
any of the proposals  \eqref{eq:proposal_1}, 
\eqref{eq:proposal_2} or \eqref{eq:proposal_3}.
The $n$-step transition is written as $P^n(x_n | x) = \mathcal{N}(x_n| \beta^n x + (1 - \beta^n) \mu, (1 - \beta^{2 n}) \Sigma)$, where $\beta = 1 - \gamma$, so that $\beta \in (-1, 1)$, and the Poisson solution from  \eqref{eq:poisson_inf_series} 
becomes
\begin{equation}
\hat{F}(x) 
= F(x) - \pi(F)  + \sum_{n=1}^{\infty} 
\left( \int \mathcal{N}(x_n| \beta^n x + (1 - \beta^n) \mu, (1 - \beta^{2 n}) \Sigma) F(x_n) d x_n  -  \pi(F) \right).
\label{eq:poisson_inf_series_gi}
\end{equation}
\end{proposition}
For several functions $F(x)$
we can obtain closed-form 
expressions of \eqref{eq:poisson_inf_series_gi}. For example, if  $F(x) = x$ we have
$$
\hat{F}(x) 
= x - \mu + \sum_{n=1}^{\infty} 
\left( \beta^n x + (1 - \beta^n) \mu  -  \mu \right) = \frac{x - \mu}{1 - \beta} = \frac{x - \mu}{\gamma},
$$
which by ignoring the constant further simplifies as
$\hat{F}(x) = \frac{x}{\gamma}$. Other exact 
closed-form solutions
are given in Table \ref{tab:solutions}.
To see that these solutions, such as $\hat{F}(x) = \frac{x}{\gamma}$, 
make the variance zero 
let us compute the MH ergodic estimator from \eqref{eq:modified_estimatorMH} by using the Gaussian invariant proposals in \eqref{eq:proposal_1}-\eqref{eq:proposal_3}. First,  given that there are no rejections and always 
$\alpha(x,y)=1$, we can simplify the estimator as
\begin{equation}
\mu_{n,G}(F) = \frac{1}{n}\sum_{i=0}^{n-1}\{F(X_i) - (G(X_i) - \E_{q(\cdot| X_i)}[G]) \},
\label{eq:estimator_optimal_Gaussianpi}
\end{equation}
where 
$\E_{q(y| X_i)}[G] =  \int G(y) q(y|X_i)dy$. For $F(x)=x$  and using
$G(x) = \hat{F}(x) = \frac{x}{\gamma}$ we have
\begin{align}
\label{eq:estimator_GIRWM_tractable}
\text{GI-RWM:} \ \  \mu_{n,G}(x) & = \frac{1}{n}\sum_{i=0}^{n-1}\left\{X_i - \frac{1}{\gamma} \left(X_i - (1- \gamma) X_i  - \gamma \mu \right) \right\}, \\
\text{GI-MALA:} \ \ \mu_{n,G}(x) 
& = 
\frac{1}{n}\sum_{i=0}^{n-1}\left\{X_i + A_{x_i} \nabla \log \pi(X_i) \right\},
\label{eq:estimator_GIMALA_tractable}
\end{align}
where we have written only 
the most advanced version of GI-MALA while the
other case is the same but with $A_{x}$  replaced by $\Sigma$. Also, we used that for the GI-RWM 
proposal in \eqref{eq:proposal_1}
it holds $\E_{q(y|X_i)}[y] = (1 - \gamma) X_i + \gamma \mu$ and for GI-MALA   $\E_{q(y|X_i)}[y] = X_i + \gamma A_{x_i} \nabla \log \pi(X_i)$.  
After some simplifications, since 
$\pi(x) = \mathcal{N}(x| \mu, 
\Sigma)$, both estimators
reduce to the exact value $\mu$
and therefore have zero variance. An advantage of 
the GI-MALA estimator
in \eqref{eq:estimator_GIMALA_tractable} versus GI-RWM 
\eqref{eq:estimator_GIRWM_tractable} is that it is somehow more black-box 
since it does not need to know $\mu$.

For functions $F(x)$ where the integrals $\int \mathcal{N}(x_n| \beta^n x + (1 - \beta^n) \mu, (1 - \beta^{2 n}) \Sigma) F(x_n) d x_n$ are tractable
but there is no closed-form of the infinite series in \eqref{eq:poisson_inf_series_gi},
we can still obtain accurate approximations by truncating the series and ignoring constants, such as  
\begin{equation}
\hat{F}(x)  \approx F(x) + \sum_{n=1}^{N}  \int \mathcal{N}(x_n| \beta^n x + (1 - \beta^n) \mu, (1 - \beta^{2 n}) \Sigma) F(x_n) d x_n,
\label{eq:poisson_inf_series_gi_tranc}
\end{equation}
where $N$ is the truncated level. 
The third column in Table \ref{tab:solutions} shows examples of such truncation-based approximate solutions.  

\begin{table}[t]
    \centering
    \caption{Exact and approximate solutions to Poisson equations.  n/n means not needed (since we have the exact solution) and n/a means not available. Also in some entries we use the shorter notation $\beta =1 - \gamma$. In the last two entries $a$ is a vector of the same size as $x$ and $b$  is a scalar, while $I(\cdot)$ denotes the indicator function and $\Phi(\cdot)$  the normal CDF.}
    \label{tab:solutions}
    \renewcommand{\arraystretch}{1.5}
    \begin{tabular}{lcc}
    \toprule 
    Function $F(x)$ & Exact Solution $\hat{F}(x)$ & Truncated Solution \\
    \midrule
      $x$   & $\frac{x}{\gamma}$ & n/n \\
      $x x^\top$ & $\frac{x x^\top  + (1-\gamma) (x \mu^\top + \mu x^\top)}{2 \gamma - \gamma^2} $ & n/n \\
      $(x -\mu) (x - \mu)^\top$ & $\frac{1}{2 \gamma - \gamma^2} (x - \mu) (x - \mu)^\top$ & n/n \\
      $e^{a^\top x}$ & n/a & $e^{a^\top x} + \sum_{n=1}^N e^{a^\top \left\{\beta^n x + (1 - \beta^n) \mu \right\}+ \frac{1}{2} (1 -\beta^{2 n}) a^\top \Sigma a}$ \\
      $I(a^\top x > b)$ & n/a & $I(a^\top x > b) + \sum_{n=1}^N \Phi\left(\frac{b - a^\top \left\{ \beta^n x + (1 - \beta^n) \mu \right\}}{\sqrt{(1 -\beta^{2 n}) a^\top \Sigma a}} \right)$ 
      \\
     \bottomrule 
    \end{tabular}
\end{table}

\subsection{Variance reduction for intractable non-Gaussian targets
\label{sec:variance_reduction}}

We now return to
the intractable non-Gaussian target case. We derive 
a sampling and variance reduction framework made of two components: (i) Collection of samples by running a  Gaussian invariant MH kernel, and (ii) construction of  the control variate estimator in \eqref{eq:modified_estimatorMH} so that the function $G(x)$ is chosen from Table \ref{tab:solutions}, i.e.,
it is an exact (or approximate) Poisson solution corresponding to the tractable Gaussian target case. 

More precisely, recall 
the ergodic estimator in \eqref{eq:modified_estimatorMH} for any non-Gaussian target $\pi$.
We assume that samples $\{X_i\}_{i=0}^{n-1}$ have been collected by running MCMC using either  the Gaussian invariant GI-RWM proposal from \eqref{eq:proposal_1} 
or GI-MALA from \eqref{eq:proposal_2}-\eqref{eq:proposal_3}. Since the target is not Gaussian,  we cannot
solve the Poisson equation to make the variance zero,  but we still hope to use a suitable function $G(x)$
that can reduce the variance.  
As $G(x)$, and for a given $F(x)$, we will use the exact (or the approximate) solution  to the Poisson equation for a Gaussian target, i.e., set $G(x) = \hat{F}(x)$ to the corresponding entry from Table \ref{tab:solutions}. 

Having specified $G(x)$,  we still need to  overcome that the estimator \eqref{eq:modified_estimatorMH}
  requires the evaluation of the  intractable integral 
$$
\int \alpha(X_i,y)(G(y)-G(X_i))q(y|X_i) d y,
$$
for each sampled point $X_i$. 
This is not feasible to compute analytically because the probability $\alpha(X_i,y)$ (as a function of $y$) can take values smaller than one and depends on the intractable target $\pi$.  
Since the MH step required one sample $Y_i \sim q(\cdot|X_i)$ when the chain was at $X_i$, an unbiased estimator of this integral can be based on the sample size one estimator $\alpha(X_i,Y_i)(G(Y_i) - G(X_i))$. This suggests that \eqref{eq:modified_estimatorMH} can be approximated by 
\begin{equation}
\label{eq:modified_estimatorMH_stoch}
\mu_{n,G}(F) = \frac{1}{n}\sum_{i=0}^{n-1}\{F(X_i) + \theta_1  \alpha(X_i,Y_i)(G(Y_i) - G(X_i))\},
\end{equation}
where we also introduced the scalar parameter $\theta_1$ in front of the control variate that we will specify shortly. 
A final trick to improve the estimator is to use a further control variate to reduce the variance of \eqref{eq:modified_estimatorMH_stoch}, by reducing the variance of the sample size one estimate $\alpha(X_i,Y_i)(F(Y_i) - F(X_i))$ of the integral  $\int \alpha(X_i,y) (F(y)-F(X_i))q(y)dy$. 
The idea, also used in 
\cite{alexopoulos2023variance}, is rather
simple. If we remove
$\alpha(X_i, Y_i)$  we can tractably 
center the function
$G(Y_i) - G(X_i)$  with respect to 
$Y_i \sim q(\cdot | X_i)$ 
according to $G(Y_i)  -  G(X_i) - \E_{q(y|X_i)} 
[G(y) - G(X_i) ] = G(Y_i) - \E_{q(\cdot|X_i)} 
[G]$. Thus, $G(Y_i) - \E_{q(\cdot|X_i)}[G]$ has zero expectation over $Y_i$ and can be used as a control variate, resulting in the estimator 
\begin{equation}
\label{eq:modified_estimatorMH_stoch2}
\mu_{n,G}(F) = \frac{1}{n}\sum_{i=0}^{n-1}\left\{F(X_i) + \theta_1 \alpha(X_i,Y_i)(G(Y_i) - G(X_i)) 
+ \theta_2 (G(Y_i) - \E_{q(\cdot|X_i)}[G])
\right\},
\end{equation}
where we introduced a second scalar parameter $\theta_2$ in front of the newly added
control variate. We observe that 
when the target $\pi$ is Gaussian so that $\alpha(X_i, Y_i)=1$ 
and by setting $\theta_1 = 1, 
\theta_2 = -1$ the overall control variate (after cancellation of 
$G(Y_i)$) reduces to  
$ - (G(X_i) -  \E_{q(\cdot|X_i)}[G])$
and thus \eqref{eq:modified_estimatorMH_stoch2} becomes the estimator in \eqref{eq:estimator_optimal_Gaussianpi}.

We select the parameters $\theta_1, \theta_2$ to minimize the asymptotic variance following \cite{dellaportas2012control}. Given that both control variates $H_1(X_i,Y_i) = \alpha(X_i,Y_i)(G(Y_i) - G(X_i))$ and $H_2(X_i,Y_i) = G(Y_i) - \E_{q(\cdot|X_i)}[G]$ in \eqref{eq:modified_estimatorMH_stoch2} are defined over the Markov chain $(X_i,Y_i)$, we choose $\theta =( \theta_1, \theta_2)$ to jointly minimize the variance of \eqref{eq:modified_estimatorMH_stoch2} over the distribution with density proportional to $\pi(x)q(y|x)$ under which $H_1$ and $H_2$ have zero mean. This gives 
\begin{equation*}
    \theta=
    \mathrm{K}_n^{-1}
    [\mu_n(FH)-
    \mu_n(F)\mu_n(H)],
\label{eq:theta}
\end{equation*}
where we defined the vector $H = [H_1, H_2]^\top \in \Re^2$, while the vectors  $\mu_n(H) = [\mu_n(H_1), \mu_n(H_2)]^\top$, $\mu_n(FH) = [\mu_n(FH_1), \mu_n(FH_2)]^\top$ are ergodic means with $\mu_n(H_j)$ and $\mu_n(FH_j)$ ($F H_j$ is a shortcut for $F(x) H_j(x,y)$) obtained from  \eqref{eq:std_estimator} and $K_n$ is a $2 \times 2$ matrix with entries $(K_n)_{ij} = \frac{1}{n-1}\sum_{t=1}^{n-1}H_i(X_t)H_j(X_t)$. 

\subsection{Algorithmic summary with an example} 
\label{sec:algo_summary} 

Here, we provide a summary of the steps needed to implement the proposed algorithms and construct a control variate  
estimator of the form in  \eqref{eq:modified_estimatorMH_stoch2}.

{\bf Step 1 (choosing the proposal).} When the target $\pi$ is non-differentiable, the only option is to use 
GI-RWM from \eqref{eq:proposal_1} where 
$(\mu, \Sigma)$  can be obtained with adaptive MCMC \citep{haario2001adaptive}. 
For differentiable targets the GI-MALA proposals in  \eqref{eq:proposal_2} and \eqref{eq:proposal_3} can be more effective and they should be preferred in practice. For such smooth differentiable targets one way to choose $(\mu,\Sigma)$ (whenever they are needed) is by applying the Laplace approximation \citep{TierneyKanade86}; see last paragraph below for more details. 

{\bf Step 2 (running the sampler).} This step is the same as in any standard MCMC algorithm such as RWM and MALA. For example, the parameter $\gamma$ is adapted to reach a desirable acceptance probability $\alpha^*$, as further described in the experiments. This is done during burn-in iterations with some standard learning rule, such as $\gamma \leftarrow \gamma (1 + \rho (\alpha(X_i,Y_i) - \alpha^*))$ 
where $\rho>0$ is a small step size. Further, to obtain the control variate
estimators with minimum post-processing cost it is convenient for the implementation of the Gaussian invariant MCMC routines to return (after burn-in) not only the samples $\{X_i\}_{i=0}^{n-1}$ (and the proposal parameters such as $\gamma, \mu,\Sigma$) but also some additional quantities calculated during the iterations. More precisely, GI-RWM returns the triples $\{X_i, Y_i, \alpha(X_i, Y_i)\}_{i=0}^{n-1}$ while GI-MALA  additionally returns the preconditioned gradient vectors $\{A_x \nabla \log \pi(X_i)\}_{i=0}^{n-1}$ 
(or $\{\Sigma \nabla \log \pi(X_i)\}_{i=0}^{n-1}$). 

{\bf Step 3 (constructing the actual estimator).} We discuss this using as an example the function $F(x)=x^{(j)}$, where $x^{(j)}$ is the $j$-component of $x$, and show how to evaluate the estimator in \eqref{eq:modified_estimatorMH_stoch2}
for both GI-RWM and GI-MALA.  Firstly, we need to choose the function $G(x)$ and according to Table \ref{tab:solutions} we select $G(x) = \frac{x^{(j)}}{\gamma}$. Then  
since for the GI-RWM proposal in \eqref{eq:proposal_1} $\E_{q(\cdot|X_i)}[G] = (1-\gamma) X_i^{(j)} + \gamma \mu^{(j)}$ the estimator \eqref{eq:modified_estimatorMH_stoch2} becomes
\begin{equation}
\frac{1}{n}\sum_{i=0}^{n-1}\left\{X_i^{(j)} + \frac{1}{\gamma} \theta_1 \alpha(X_i,Y_i)(Y_i^{(j)} - X_i^{(j)}) 
+ \frac{1}{\gamma} \theta_2 (Y_i^{(j)} - (1-\gamma) X_i^{(j)} - \gamma \mu^{(j)})
\right\},
\label{eq:estimator_gi_rwm_x}
\end{equation}
where the coefficients $(\theta_1, \theta_2)$
are obtained by solving the $2 \times 2$ linear system described in Section \ref{sec:variance_reduction}\footnote{Note that for each different $j$-th dimension $x^{(j)}$ of $x$ we estimate a separate pair $(\theta_1,\theta_2)$.}. 
For GI-MALA with preconditioning 
$A_x$ (or $\Sigma$) observe that 
$\E_{q(\cdot|X_i)}[G] 
= X_i^{(j)} + \gamma [A_{x_i}  \nabla \log\pi(X_i))]_j$ and thus the estimator becomes
\begin{equation}
\frac{1}{n}\sum_{i=0}^{n-1}\left\{X_i^{(j)} + \frac{1}{\gamma}\theta_1  \alpha(X_i,Y_i)(Y_i^{(j)} - X_i^{(j)}) 
+ \frac{1}{\gamma} \theta_2 (Y_i^{(j)} - X_i^{(j)} - \gamma [A_{x_i}  \nabla \log\pi(X_i))]_j
\right\},
\label{eq:estimator_gi_mala_x}
\end{equation}
where  
$(\theta_1, \theta_2)$ are 
estimated as described in 
Section \ref{sec:variance_reduction}.
These estimators are used for intractable non-Gaussian targets. However, as a sanity check, observe that
when the target is  $\mathcal{N}(x|\mu, \Sigma)$,  the acceptance probabilities are always $\alpha(X_i, Y_i)=1$ and by setting the values $\theta_1 = 1, 
\theta_2 = -1$, the above estimators reduce 
to the zero variance estimators in \eqref{eq:estimator_GIRWM_tractable}-\eqref{eq:estimator_GIMALA_tractable}. 

Finally, 
for more complex functions
such as $F(x) = e^{a^\top x}$
the $G(x)$ function in Table \ref{tab:solutions}
can depend on $(\mu, \Sigma)$.
When the sampler itself (step 2 above) does not output values for $(\mu,\Sigma)$, as  for the most advanced GI-MALA proposal in 
\eqref{eq:proposal_3}, 
then to evaluate the estimator (step 3 above) we consider
the Laplace approximation 
\citep{TierneyKanade86}
to locate the mode $m^* = \text{argmax}_x \{\log \pi(x) \}$ and set $\mu= m^*$,$\Sigma = - \left[ \nabla^2 \log \pi(m^*) 
\right]^{-1}$.
Another option 
could be to run a separate Markov chain and then use Monte Carlo estimates for $(\mu,\Sigma)$.

\section{Application to latent Gaussian models}
\label{sec:MCMC_GPs}

As a main application  we consider latent Gaussian models which have the form  
\begin{equation}
\label{eq:LGM}
\pi(x) \propto \exp\{g(x)\}\dN(x|0,\Sigma_0).
\end{equation}
Latent Gaussian models range from Gaussian processes and 
Markov random fields \citep{rasmussen2006gaussian, Neal99,Rueetal2009} to Bayesian inverse problems \citep{cotter2013mcmc, beskos2017geometric}. In these models, it is typically assumed that the observations given the latent state $x$ are independent and thus $g(x) = \sum_{i=1}^d g(x^{(i)})$. To construct a sampler we rely on the GI-MALA proposal from \eqref{eq:proposal_3} where we specify a fast 
approximation $A_x$ to the negative inverse Hessian matrix. The exact negative Hessian is 
$$
- \nabla^2 \log \pi(x) =  -\nabla^2 \log\dN(x|0,\Sigma_0)  - \nabla^2 g(x) = \Sigma_0^{-1} + L_x,
$$
where $L_x$ is a diagonal 
matrix with 
entries 
$(L_x)_{ii} = -\frac{\partial^2 g(x)}{\partial x^{(i)}\partial x^{(i)}},\,\,\, i=1,\ldots,d$.
One option is to set $A_x$ to the exact negative inverse Hessian $(\Sigma_0^{-1} + L_x)^{-1}$. 
However, this can be very expensive since if the 
diagonal matrix 
$L_x$ is non-isotropic (i.e., it is not a scalar times identity), then at each MCMC iteration we need to apply an $O(d^3)$ decomposition of $\Sigma_0^{-1} + L_x$.
In contrast, 
current samplers 
that preserve Gaussian invariance 
to the prior 
$\mathcal{N}(x|0,\Sigma_0)$, such as pCN  
\citep{beskos2008mcmc,cotter2013mcmc, Neal99} and the  methods in \cite{titsias2018auxiliary}, have $O(d^2)$ cost per MCMC iteration.
Thus, to be as efficient as these methods we will 
build a fast $O(d^2)$
approximation to $(\Sigma_0^{-1} + L_x)^{-1}$.

Without loss of generality 
we will assume that the negative second
derivative matrix of log likelihood 
$- \nabla^2 g(x)$
does not contain an additive
constant term. 
If it were  $- \nabla^2 g(x) = - \nabla^2 \tilde{g}(x) + C$ with 
$C$ a constant diagonal matrix,  then we could absorb $C$ into 
the Gaussian prior 
covariance by redefining the Gaussian prior to have  
covariance  $\Sigma_0 \leftarrow (\Sigma_0^{-1} + C)^{-1}$ and using a new likelihood proportional to  $\exp\{\tilde{g}(x)\}$. Under this assumption, 
at each MCMC iteration 
we introduce as a preconditioning the matrix 
\begin{equation}
A_x = (\Sigma_0^{-1} + \delta_x I_d)^{-1}, \ \ 
\delta_x = \text{mean}\{ (L_x)_{11}, \ldots, (L_x)_{dd} \},
\label{eq:preconditioning_latentGaussian}
 \end{equation}
 where $\delta_x$ 
 is the mean of all diagonal entries of $L_x$. Thus, 
 the proposal in \eqref{eq:proposal_3} takes the form 
\begin{equation}
    \label{eq:GP_prop}
    q(y | x) = \dN(y | x + \gamma A_x(\nabla g(x) - \Sigma_0^{-1} x), (2 \gamma - \gamma^2) A_x ),
\end{equation}
where we used that 
$\nabla \log \pi(x) = \nabla g(x) - \Sigma_0^{-1} x$. The following 
Proposition \ref{prop:GP_prop_properties} shows how the MCMC algorithm with proposal 
\eqref{eq:GP_prop} 
can be implemented with $O(d^2)$ cost per iteration.

\begin{proposition}
\label{prop:GP_prop_properties}
Let $\pi(x) \propto \exp\{g(x)\}\dN(x|0,\Sigma_0)$, $x \in \Re^d$, 
and $\Sigma_0 = U \Lambda U^\top$ 
is a precomputed eigenvalue decomposition of the prior covariance matrix $\Sigma_0$ with $\lambda_i = (\Lambda)_{ii}$ denoting an eigenvalue. Each MH iteration with proposal \eqref{eq:GP_prop} can be implemented with $O(d^2)$ cost as 
\begin{itemize}
    \item[(i)] Given $\zeta_x = U^\top (x + \delta_x^{-1} \nabla g(x))$
and $\epsilon \sim \mathcal{N}(0,I_d)$, sampling from  $q(y|x)$ is done as
    $$
y = (1-\gamma) x  
+  U  \left[ \gamma \Lambda (\Lambda + \delta_x^{-1} I  )^{-1} \zeta_x + \sqrt{\delta_x^{-1} (2 \gamma - \gamma^2)} \Lambda^{\frac{1}{2}} (\Lambda  + \delta_x^{-1} I )^{-\frac{1}{2}} 
\epsilon \right].   
$$ 
\item[(ii)] The MH probability is computed as
    $
      a(x,y) = \text{min}\{1, e^{g(y) - g(x) + h(x,y) - h(y,x)}\},
    $
    where
    \begin{equation}
    \label{eq:hxy}
h(x,y) 
 = \frac{1}{2} \frac{\delta_x}{(2\gamma - \gamma^2)} ||y - x - \frac{\gamma}{\delta_x} \nabla g(x) ||^2 -\frac{1}{2} \sum_{i=1}^d \log (\lambda_i \delta_x + 1)  - \frac{1}{2}\frac{\gamma}{(2 - \gamma)} \zeta_x^\top \left(\Lambda + \delta_x^{-1} I_d \right)^{-1} \zeta_x.
 \nonumber
\end{equation}
\end{itemize}
\end{proposition}
The proof is given in the appendix. 
Note that the assumption we made that the log-likelihood 
$g(x)$ has no constant additive terms in the second derivatives and if it has  then this is absorbed into
$\Sigma_0$, 
guarantees that the 
preconditioning $A_x$ in 
\eqref{eq:preconditioning_latentGaussian} is Gaussian optimal (see Definition \ref{def:preconditioning}). This means that for conjugate 
latent Gaussian models, such
as Gaussian process regression with log-likelihood $g(x) = - \frac{1}{2 \sigma^2} ||y - x||^2 + const$, 
the proposal from \eqref{eq:GP_prop} is reversible to the target and in fact for $\gamma=1$ gives i.i.d.\ sampling. In such case we can also obtain zero variance estimation
by using exact solutions to Poisson equations.  

The ergodic estimators for the above GI-MALA sampler follow the general form in
\eqref{eq:modified_estimatorMH_stoch2}. For instance, if $F(x)=x$ the estimator is 
\eqref{eq:estimator_gi_mala_x}.

\section{Experiments
\label{sec:experiments}
} 

Sections \ref{sec:ess}  
and \ref{sec:variancereductionresults}
consider several examples in Bayesian logistic regression and high-dimensional latent Gaussian models.  Specifically, Section \ref{sec:ess} 
compares the GI-MALA sampler with previous algorithms in terms of effective sample size (ESS) only, i.e., without incorporating variance reduction. Section \ref{sec:variancereductionresults} 
shows how the Gaussian invariant estimators
can benefit from the use of control 
variates and the analytical solution to the Poisson equation for the function 
$F(x)=x$. 
Finally, Section 
\ref{sec:t-example}
demonstrates the ability 
to reduce the variance when estimating the expectation of the 
function $F(x) = I(a^\top x > b)$, where from Table \ref{tab:solutions} only an approximate truncated solution exists.

All computations were performed on a Mac Studio workstation equipped with an Apple M2 Ultra SoC (24-core CPU), 64 GB unified memory, running macOS 26.1. 
Replication code in MATLAB can be found in \url{https://github.com/angelosalexopoulos/gi_mala}.

\subsection{Results without the use of control variates}
\label{sec:ess}

We consider medium and high dimensional problems in Bayesian logistic regression (Section 
\ref{sec:logistic})
and latent Gaussian models (Sections \ref{sec:gp_classification} 
and \ref{sec:cox_processs}).
In Bayesian logistic regression we 
compare GI-MALA from 
(\ref{eq:proposal_2}) 
with MALA 
from (\ref{eq:MALA}) where both have the same constant preconditioning $\Sigma$.   
For the latent Gaussian models
we consider position-dependent preconditioning 
and we compare GI-MALA from  
(\ref{eq:proposal_3}) with the 
corresponding MALA 
that has the same preconditioning and drift as GI-MALA, but it scales
the covariance
with $2 \gamma$:  
\begin{equation}
    \label{eq:MALA_experiments}
q_{MALA}(y|x) = \dN(y|x + \gamma A_x \nabla \log \pi(x), 2 \gamma A_x).
\end{equation}
For latent Gaussian models we also 
include in the comparison 
several other samplers reported in \cite{titsias2018auxiliary}, such as the so called marginal sampler \citep{titsias2018auxiliary}, 
as well as earlier algorithms such as pCN, pCNL \citep{cotter2013mcmc} and Elliptical slice sampling \citep{nishihara2014parallel}. Note 
that we do not include in the comparison
GI-RWM since this is equivalent to 
pCN.

All the comparisons are made with respect to ESS also taking into account the running times of the different algorithms which all are tuned to achieve the acceptance rate described as optimal in the original papers. For instance, for the MALA proposal in 
(\ref{eq:MALA}) or 
(\ref{eq:MALA_experiments})
the target acceptance 
rate is $0.574$ as suggested by \cite{roberts2001optimal}. 
For the proposed GI-MALA, 
we follow the guidelines  from the optimal scaling analysis in Section \ref{sec:optimal_scaling} 
where the optimal acceptance rate for GI-MALA is shown to be target-dependent and typically higher than 
$0.574$ of MALA. Therefore, 
we tune the constant $\gamma$ in GI-MALA to achieve an acceptance rate between $75\%$ and $85\%$. The adaptation phase for all samplers takes place only during burn-in iterations, while at collection of samples stage the step size $\gamma$ is fixed.

\subsubsection{Bayesian logistic regression
\label{sec:logistic}
}

We use several Bayesian logistic regression datasets 
as also considered by \cite{girolami2011riemann,titsias2019gradient,alexopoulos2023variance}. These consist of an $N$-dimensional vector of binary responses and an $N \times d$ matrix with covariates including a column of ones. We consider a Bayesian logistic regression model with an improper prior for the regression coefficients $w \in \Re^d$ of the form $p(w) \propto 1$. We draw samples from the posterior distribution of $w$ by employing the proposed GI-MALA sampler  as well as MALA where we set the preconditioning $\Sigma$ based on the Laplace approximation around the mode $w^*$. Both samplers are initialized at $w^*$ and run for $5000$ burn-in iterations and subsequently $10^4$ posterior samples were collected. Table \ref{tab:logistic_low} reports ESS for three datasets having different dimension $d$ and sample size $N$. In all cases  GI-MALA outperforms MALA. In particular, the ESS for GI-MALA is $1.5$ to $3$ times larger than the ESS of MALA.

\begin{table}[H]
\caption{ESS scores in Bayesian logistic regression with $d$-dimensional posterior distributions of datasets with $N$ examples.
All numbers are averages across ten random repeats.
}
 \centering
\begin{tabular}{lllll}
\toprule
Dataset & Method & Min & Median & Max \\
\midrule
Heart ($N=270, d=13$) & GI-MALA & 2751.3 & 3524.9 & 4010.6 \\
         & MALA   & 1992.1 & 2238.1 & 2460.2 \\
Australian ($N=690, d=14$) & GI-MALA & 3818.5 & 4681.4 & 5278.2 \\
         & MALA   & 2031.7 & 2247.3 & 2486.6 \\
German ($N=1000, d=24$) & GI-MALA & 3465.2 & 5380.2 & 6130.9 \\
         & MALA   & 1550.0 & 1831.9 & 1998.1 \\
\bottomrule
\end{tabular}
\label{tab:logistic_low}
\end{table}

\subsubsection{Binary Gaussian process classification}
\label{sec:gp_classification}

We consider binary Gaussian process classification where $\{y_i, z_i\}_{i=1}^d$ denotes a set of binary labels $y_i$ and input vectors $z_i$. The binary labels $y_i$ given latent function values $x_i := x(z_i)$ are drawn independently  from a logistic regression likelihood such that $g(x)$ in \eqref{eq:LGM} is
$g(x) = \sum_{i=1}^d y_i \log \sigma(x_i) + (1 - y_i) \log (1 - \sigma(x_i))$,
where $\sigma(x_i) = \frac{1}{1 + \exp(-x_i)}$ is the logistic function. The logit values 
$x:=(x_1,\ldots, x_d)$ follow a Gaussian process prior, so that $x \sim \mathcal{N}(0, \Sigma_0)$ and where $\Sigma_0$ is defined by the squared exponential covariance function $k(z_i,z_j) = \sigma_x^2 \exp\{ - \frac{1}{2 \ell^2} ||z_i - z_j||^2 \}$. We compare the samplers in five standard binary classification datasets \citep{girolami2011riemann, titsias2018auxiliary, alexopoulos2023variance}. This involves 
applying MCMC in latent Gaussian models with dimensionality ranging from $d=250$ to $d=1000$. Here
we show results for the Heart and Australian Credit datasets for which the latent dimension is $d=270$ and $d=690$, while the remaining results are in the appendix. For all 5 datasets the samplers run for $10^4$ iterations from which the first $5000$ is the adaptation burn-in phase.
All samplers are initialized at $x=0$.
The ESS and running times for all methods and the two datasets are presented in Tables \ref{tab:heart} and \ref{tab:australian}. It is clear that GI-MALA has the greatest ESS as well as the highest overall efficiency score (see Min ESS/s).

\begin{table}[H]
\caption{Comparison of sampling methods in Heart dataset. The size of the latent vector
$x$ is $d = 270$. All numbers are averages across ten random repeats. In parentheses we report the standard deviation of the minimum ESS/s.}
\centering
\begin{tabular}{lllll}
\toprule
\em Method &\em Time(s)  &\em ESS (Min, Med, Max)  &\em Min ESS/s  \\ 
\midrule 
GI-MALA  &  0.6  & (319.7, 859.1, 1446.7)  &  570.04 (73.73)\\ 
MALA  &  0.8  &  (129.3, 309.5, 513.9)  &  169.12 (23.57)\\ 
mGrad  &  0.5  &  (192.1, 599.9, 1067.2)  &  363.88 (76.10)\\ 
Ellipt  &  1.0  &  (12.6, 43.1, 96.5)  &  12.34 (1.47)\\ 
pCN  &  0.3  & (8.2, 28.6, 75.6)  &  24.93 (4.13)\\ 
pCNL  &  0.4  &  (22.8, 69.5, 144.5)  &  53.36 (10.12)\\ 
\bottomrule
\end{tabular}
\label{tab:heart}
\end{table}

\begin{table}[H]
\caption{Comparison of sampling methods in Australian Credit dataset. The size of the latent vector
$x$ is $d = 690$. All numbers are averages across ten random repeats. In parentheses we report the standard deviation of the minimum ESS/s.}
\centering
\begin{tabular}{llll}
\toprule
\em Method &\em Time(s)  &\em ESS (Min, Med, Max)  &\em Min ESS/s  \\ 
\midrule
GI-MALA  &  1.3  & (294.6, 714.8, 1552.0)  &  226.23 (17.62)\\ 
MALA  &  1.8  & (86.5, 213.1, 440.5)  &  49.13 (8.45)\\ 
mGrad  &  1.2  & (206.0, 502.0, 1201.9)  &  165.78 (14.11)\\ 
Ellipt  &  2.3  & (6.0, 25.1, 97.8)  &  2.65 (0.31)\\ 
pCN  &  0.7  &  (5.1, 17.8, 73.0)  &  7.25 (0.77)\\ 
pCNL  &  1.1  &  (5.4, 21.2, 88.0)  &  4.88 (0.51)\\ 
\bottomrule
\end{tabular}
\label{tab:australian}
\end{table}

\subsubsection{Log-Gaussian Cox process}
\label{sec:cox_processs}

Following \cite{girolami2011riemann}  
we consider a very high dimensional log-Gaussian Cox model where an area of interest $[0,1]^2$ is discretized into a $64 \times 64$ regular grid, and each latent variable $x_{ij}$ is associated with the grid cell $(i,j),i,j,=1,\ldots,64$. 
The dimension of $x$ is $d=4096$. Then, a data vector of counts is generated independently given a latent intensity process $\lambda(\cdot)$. Each $y_{ij}$ follows a Poisson distribution with mean $\lambda_{ij} = m \exp(x_{ij} + v)$, where $m = 1/4096$ represents the area of the grid cell and $v$ is a constant mean value. 
The latent vector $x$ is drawn from a Gaussian process with zero mean and covariance function
$
k((i,j),(i',j')) = \sigma_x^2 e^{-\frac{\sqrt{(i - i')^2 + (j - j')^2}}{64 \beta}}.
$  
The likelihood function is of Poisson form, so
$
g(x) = \sum_{i,j}^{64} \left( y_{ij} (x_{ij} + v) - m \exp(x_{ij} + v) \right).
$ 
We utilized the dataset from \cite{girolami2011riemann}, which was simulated using the above model with hyperparameters set to $\beta = 1/33$, $\sigma_x^2 = 1.91$, and $v = \log(126) - \sigma_x^2/2$. Our objective was to infer the latent vector $x$ while holding the hyperparameters $(\beta, \sigma^2, v)$ fixed at their true values.  
We follow \cite{girolami2011riemann} and \cite{titsias2018auxiliary} and run the algorithms for 5000 burn-in iterations and subsequently collect 5000 posterior samples. 

Further, in 
this example we also include in the comparison the Riemann manifold Hamiltonian Monte Carlo (RHMC) of  \cite{girolami2011riemann} which runs in $O(L d^2)$ time per iteration where $L$ is the number of leap-frog steps. Table \ref{tab:logGaussianCox} compares the different samplers.  We observe that  GI-MALA has the highest  Min ESS/s. The second best is mGrad \citep{titsias2018auxiliary}. Interestingly, the third best is RHMC  \citep{girolami2011riemann} which has better ESS than both GI-MALA and mGrad but its computational cost is  larger than those algorithms. This is because 
RHMC performs $L=30$ leap-frog steps per iteration, which requires multiple gradient evaluations per iteration and results in lower overall efficiency (Min ESS/s) than the other two methods.

\begin{table}[H]
\caption{Comparison of sampling methods in the log-Gaussian Cox model dataset where $d=4096$. All numbers are averages across
ten random repeats.}
\centering
\begin{tabular}{lllll}
\toprule
\em Method &\em Time(s)  &\em ESS (Min, Med, Max)  &\em Min ESS/s \\ 
\midrule
GI-MALA  &  100.9  &  (276.7, 1010.5, 1753.2)  &  2.74 (0.39)\\ 
MALA  &  148.3 & (34.3, 125.4, 249.1)  &  0.23 (0.07)\\ 
mGrad  &  124.3  & (168.4, 800.3, 1687.2)  &  1.36 (0.39)\\ 
Ellipt  &  29.5  & (4.6, 17.4, 67.2)  &  0.16 (0.01)\\ 
pCN  &  25.7  & (4.0, 12.0, 54.0)  &  0.15 (0.01)\\ 
pCNL  &  99.1  & (4.0, 12.6, 60.2)  &  0.04 (0.00)\\ 
RHMC  &  1204.2  &  (1909.2, 4580.6, 5000.0)  &  1.59 (0.10)\\ 
\bottomrule
\end{tabular}
\label{tab:logGaussianCox}
\end{table}

\subsection{Variance reduction results
\label{sec:variancereductionresults}}

Here, we study the effect 
of incorporating the control variates in GI-MALA. Using the same samples obtained by GI-MALA, we compare the variance  of the standard ergodic mean estimator $\mu_{n}(F)$ in \eqref{eq:std_estimator} with the variance of the modified estimator $\mu_{n,G}(F)$ given by \eqref{eq:estimator_gi_mala_x} for $F(x) = x$. To compute the
variance  we repeat the MCMC experiments and the computation of the estimators
100 times under different random seeds. More precisely, to estimate the variance of $\mu_{n}(F)$ we obtained $T=100$ different  estimates $\mu_{n}^{(i)}(F)$, $i=1,\ldots,T$, for $\mu_{n}(F)$ based on $T$ independent MCMC runs. Then, the variance of $\mu_{n}(F)$ is estimated by 
\begin{equation}
\label{eq:variance_estimator}   \frac{1}{T-1}\sum_{i=1}^{T}\{\mu_{n}^{(i)}(F)-\bar {\mu}_{n}(F)\}^2,
\end{equation}
where $\bar {\mu}_{n}(F)$ is the average.  
Similarly, we estimate the variance of $\mu_{n,G}(F)$.

In Tables \ref{tab:logistic_var}, \ref{tab:GP_logistic_var} and \ref{tab:GP_poisson_var} we consider all target distributions and the exact MCMC settings from Section \ref{sec:ess} and we report the variance ratios of the standard ergodic estimator $\mu_{n}(F)$ and the enhanced estimator $\mu_{n,G}(F)$ with the control variates. As we can observe by using the enhanced estimator we can often achieve significant variance reduction.  
Further results for the remaining datasets are given in the appendix.

\begin{table}[H]
\centering
\caption{
Range of estimated factors (variance ratios) by which the variance of $\mu_n(F)$ is larger than the variance of $\mu_{n,G}(F)$ for the Bayesian logistic regression targets. Each entry in the table shows the smallest variance factor (on the left of $-$)
and the largest variance factor (on the right of $-$) across all $d$ dimensions of the state vector $x$.}
\begin{tabular}{llllll}
\toprule
  Dataset  & $n=1,000$& $n=10,000$ & $n=50,000$ & $n=200,000$     &  \\
\midrule
Heart ($d=13$) & 2.17-3.20 & 1.97-3.48 & 2.06-2.90 & 2.22-2.79 \\
Australian ($d=14$) & 2.33-3.53 & 2.40-3.81 & 1.75-3.24 & 1.81-3.99 \\
\bottomrule
\end{tabular}
\label{tab:logistic_var}
\end{table}

\begin{table}[H]
\centering
\caption{Estimated factors for the Gaussian process binary classification targets.}
\begin{tabular}{llllll}
\toprule
  Dataset  & $n=1,000$& $n=10,000$ & $n=50,000$ & $n=200,000$     &  \\
\midrule
Heart ($d=270$) & 1.90-5.00 & 1.81-5.13 & 1.89-5.68 & 1.88-5.41 \\
Australian ($d=690$) & 1.77-4.44 & 1.89-4.76 & 1.94-4.56 & 1.92-4.67 \\
\bottomrule
\end{tabular}
\label{tab:GP_logistic_var}
\end{table}

\begin{table}[H]
\centering
\caption{Estimated factors for the Gaussian process log-Cox model with $d=4096$.}
\begin{tabular}{lll}
\toprule
  $n=1,000$& $n=10,000$  &  \\
\midrule
1.59-5.99 & 1.59-5.23 \\
\bottomrule
\end{tabular}
\label{tab:GP_poisson_var}
\end{table}

\subsubsection{An example using the truncated solution} 
\label{sec:t-example}

\begin{figure}[H]
    \centering
    \includegraphics[width=0.95\linewidth]{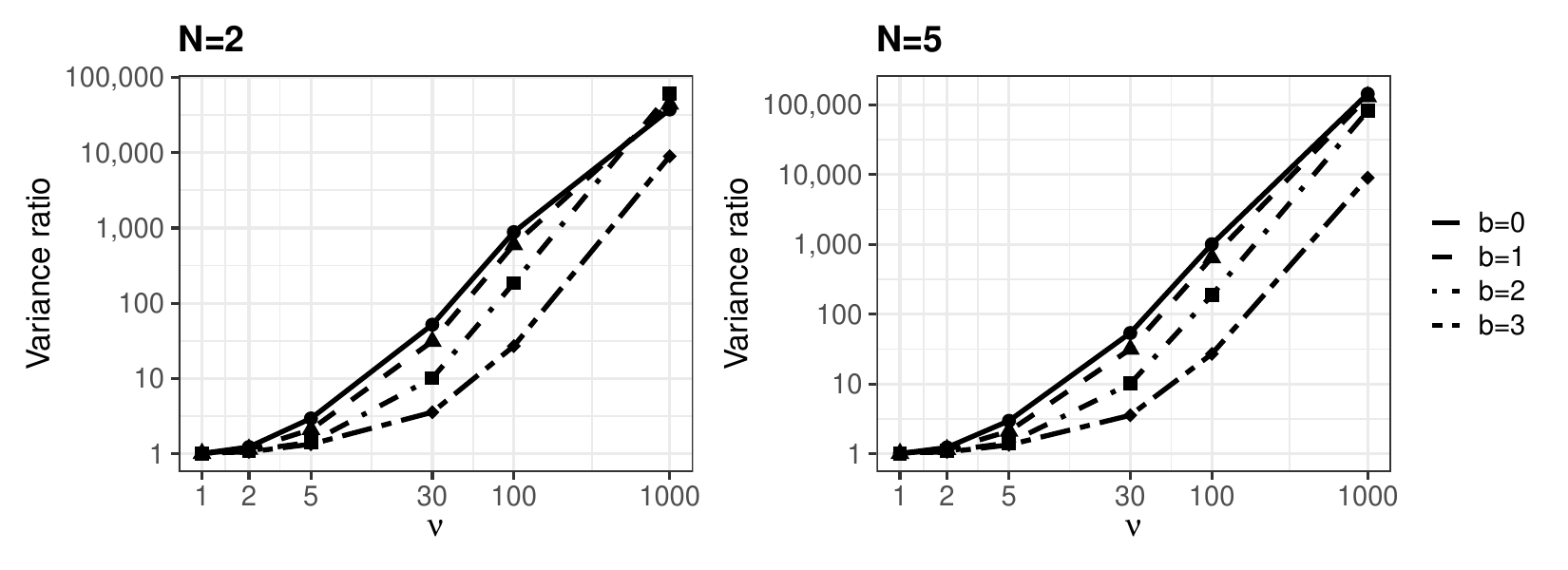}
    \caption{Estimated factors (variance ratios) by which the variance of $\mu_{n}(F)$ is larger than the variance of $\mu_{n,G}(F)$ when the target is a univariate Student's t-distribution with $\nu$ degrees of freedom and $F(x)= I(x > b)$, for different values of $\nu$, $b$ and $N$.}
    \label{fig:var_ratio_t}
\end{figure}

To illustrate the estimators on a more complex $F$, 
here we  estimate tail probabilities where $F(x)= I(a^\top x > b)$ and the corresponding $G(x)$
from Table 1 is the   truncated solution  
\begin{equation}
\label{eq:approxG}
G(x) = I(a^\top x > b) + \sum_{n=1}^N \Phi\left(\frac{a^\top \left\{ \beta^n x + (1 - \beta^n) \mu \right\}- b}{\sqrt{(1 -\beta^{2 n}) a^\top \Sigma a}} \right), 
\end{equation}
where $(\mu, \Sigma)$ are specified 
by running the Laplace approximation to locate the mode $m^*$
of $\log \pi$ and then setting $\mu=m^*$ and $\Sigma = - \left[\nabla^2 \log \pi(m^*) \right]^{-1}$. 
We take the target to be the univariate Student's $t$-distribution with density 
$$
\pi(x|\nu) = \frac {\Gamma \left({\frac {\ \nu +1\ }{2}}\right)}{{\sqrt {\pi \ \nu \ }}\ \Gamma \left({\frac {\nu }{\ 2\ }}\right)}\ \left(\ 1+{\frac {~x^{2}\ }{\nu }}\ \right)^{-{\frac {\ \nu +1\ }{2}}}.
$$
Observe that the mode is $\mu=0$ 
and $\Sigma = \frac{\nu}{\nu+1}$. 
Then we run the GI-MALA proposal from \eqref{eq:proposal_2} with the above preconditioning $\Sigma = \frac{\nu}{\nu +1}$, to draw samples from the target.  We aim to estimate $\pi(F)$ for $F(x)= I(a x > b)$ where  we set $a =1$ and consider different values for $b$.
Figure \ref{fig:var_ratio_t} compares the variance of the standard ergodic mean estimator $\mu_{n}(F)$ in \eqref{eq:std_estimator} with the variance of the modified estimator $\mu_{n,G}(F)$, given by \eqref{eq:estimator_gi_mala_x}, across different values of $\nu,b$ and truncation level $N$; see also the appendix for a related Table.

From the results we observe that when the target is very different from Gaussian (degrees  of freedom $\nu=1,2$) the variance reduction is small. As the degrees of freedom increase we 
obtain substantial variance reduction. 
Also, note that the truncation level $N=2$ 
is already sufficient 
for the function $G(x)$ to define a useful control variate, while by increasing the level  
to $N=5$ we only observe
a minor improvement 
for the large degrees of freedom. Finally, the increasing difficulty of the function $F(x) = I(x>b)$, as $b$ gets larger, also affects the variance reduction factors so that for larger $b$ the variance reduction is smaller.

\section{Theory
\label{sec:theory}
}

 We provide precise conditions under which our algorithm is geometrically ergodic and derive convergence rates for the truncated approximation \eqref{eq:poisson_inf_series_gi_tranc} to the solution of the Poisson equation.  We also establish that our final estimators follow the law of large numbers (LLN) and the central limit theorem (CLT).  Finally, we study the optimal acceptance rate and the choice of $\gamma$ for a finite dimension $d$. We follow the developments of  \cite{roberts1996exponential,roberts1996geometric} who provided the uniform geometric ergodicity of MALA under certain restrictive conditions. This theory was later extended by \cite{roy2023convergence} to address position-dependent versions of MALA. Furthermore, recent work by \cite{oliviero2024geometric} provides a quantitative analysis of convergence rates for MALA. To establish LLN and CLT, we directly follow the theory developed by \cite{dellaportas2012control}.  The works of  
 \cite{gelman1997weak,roberts1998optimal,roberts2001optimal} and \cite{titsias2019gradient} inspired the development of the optimal scaling results.

All proofs can be found in the appendix. 

\subsection{Geometric ergodicity}
Let $\Vert x \Vert$ denote the Euclidean norm of a vector $x\in\Re^d$  and 
$\Vert A \Vert:=\sup_{x\in\Re^d,x \neq 0}\{\Vert Ax\Vert/\Vert x\Vert\}$ denote the matrix norm of a matrix  $A\in \Re^{d\times d}$. 
For a measurable Borel function $f$ and some function $V: \Re^d \rightarrow [1,\infty)$, the $V$-norm of $f$ is given by $\Vert f \Vert_V:=\sup_{x\in \Re^d} |f(x)|/V(x)$.  We define the function space
$
    L_V^\infty := \{ f: \Vert f\Vert_V<\infty\}
$
and the $V$-norm distance 
$
\Vert P-Q\Vert_V := \sup_{f\in L_V^\infty}|Pf(x)-Qf(x)|
$ 
for some measures $P$ and  $Q$. 
Moreover,  denote by $R_A(x):=\{y:\pi(x)q(y|x)\leq \pi(y)q(x|y)\}$ the acceptance region of the GI-MALA algorithm and by  $R_R(x):=R_A(x)^c$ the potential rejection region. Finally, define $I(x):=\{y:\Vert y\Vert \leq \Vert x\Vert\}$, $\mu(x)=x+\gamma A_x \nabla \log\pi(x)$ and $\Sigma_x = (2\gamma-\gamma^2)A_x$. 
The following assumptions are convenient for the proof of the geometric ergodicity of our algorithm, and they have been discussed in some detail in \cite{roberts1996exponential,roberts1996geometric} and \cite{roy2023convergence}.
\begin{assumption}[A1]\label{thm:A1}
For all x, there exists a constant $k > 0$, such that $s:=2k \sup_x \Vert \Sigma_x\Vert <1$ and  
$ \lim_{\Vert x\Vert \rightarrow \infty} \inf \left\{\Vert x\Vert^2 - \frac{\Vert\mu(x)\Vert^2}{1-s}\right\} > \frac{d}{2k}\log\frac{1}{1-s}.
$
Moreover, $\mu(x)$ is bounded on bounded sets.
    
\end{assumption}
\begin{assumption}[A2]\label{thm:A2}
    The acceptance region $R_A(x)$ converges inward in the proposal $q(y|x)$:  $$ \lim_{\Vert x\Vert\rightarrow\infty}\int_{R_A(x)\Delta I(x)}q(y|x)dy = 0
    $$
    where $R_A(x)\Delta I(x) = (R_A(x)\cap I(x))/(R_A(x)\cup I(x))$ is the symmetric difference.
\end{assumption}

\begin{theorem}\label{thm:ergodicity}
    Under (A\ref{thm:A1}),  (A\ref{thm:A2}) and $V(x) = \exp\{kx^\top x\}$,  the transition kernel \eqref{eq:transition_MH} with the proposal \eqref{eq:proposal_3} is $V$-uniform geometrically ergodic, i.e., there exist constants $C<\infty$ and $\rho \in (0,1)$ such that 
    $$
        \Vert P(y|x)^N-\pi(x)\Vert_V \leq C\rho^N V(x), \ \forall N.
    $$
\end{theorem}

\subsection{Truncated Poisson solutions}

For Gaussian targets and GI-MCMC proposals we use the truncated Poisson solution from \eqref{eq:poisson_inf_series_gi_tranc}, which has an approximation or truncation  error. This motivates us 
to study this truncation error not only for the case in \eqref{eq:poisson_inf_series_gi_tranc}, but more broadly for truncated solutions of any MCMC transition kernel. In Theorem \ref{thm:converge_sol} we derive a general error bound assuming a generic target $\pi$ and $\pi$-invariant transition kernel $P$. 
 Then, in Proposition \ref{thm:gaussian_target} 
 we specialize this bound
 to Gaussian targets and the specific truncated solution from
\eqref{eq:poisson_inf_series_gi_tranc}.

We will make use of the following assumption.


\begin{assumption}[A3]\label{thm:A3}
 $F\in L^\infty_V$, $\pi(V) <\infty$.
\end{assumption}
For a general target $\pi$
and transition kernel $P$ we truncate the Poisson solution in \eqref{eq:poisson_inf_series}
as 
$$
        \hat{F}_N = F-\pi(F) + \sum_{n=1}^{N} (P^n F-\pi(F))
$$
and we define the truncation error
$
    \Delta_{N} := \Big|P \hat{F}_N - \hat{F}_N - (-F + \pi(F))\Big|.
$
\begin{theorem}\label{thm:converge_sol}
    Under (A\ref{thm:A1}), (A\ref{thm:A2}) and (A\ref{thm:A3}) there exist constants $C'<\infty$ and $\rho \in [0,1)$ such that
    $
        \Delta_N \leq C'\rho^N V(x).
    $
\end{theorem}
Theorem \ref{thm:converge_sol} provides an error bound for the truncated solution to the Poisson equation for any target,  
 transition kernel and function $F$ satisfying (A\ref{thm:A1}), (A\ref{thm:A2}) and (A\ref{thm:A3}). The error $\Delta_{N}$ converges geometrically at rate $\rho$. Next we specialize this result to a Gaussian target with Gaussian invariant proposal, where we can more  explicitly evaluate the truncation error.

\begin{proposition}\label{thm:gaussian_target}
    When the target is Gaussian $\mathcal{N}(x|\mu,\Sigma)$ and we apply MH using
any of the proposals  \eqref{eq:proposal_1}, 
\eqref{eq:proposal_2} or \eqref{eq:proposal_3}, the truncated solution is given by \eqref{eq:poisson_inf_series_gi_tranc}. Assuming $\pi( |F(y)| \cdot\Vert(y-\mu)^\top \Sigma^{-1}\Vert)<\infty$, we obtain
    $$
        \Delta_N \leq B|\beta|^{N} \Vert x-\mu\Vert + O(\beta^{2N})
    $$ where $\beta \in (0,1)$ and $B = |\beta|\int |F(y)|\cdot \Vert(y-\mu)^\top \Sigma^{-1}\Vert \mathcal{N}(y|\mu,\Sigma)dy$.
\end{proposition}
In Proposition \ref{thm:gaussian_target}, the convergence rate $\rho$ corresponds to $|\beta|$, the resulting bound depends linearly on $\Vert x-\mu \Vert$ up to a higher-order remainder, and the optimal step size is $\beta = 0$ or $\gamma = 1$, corresponding to the immediate convergence of the solution to the Poisson equation.

\subsection{LLN and CLT for the estimators}
Theorem \ref{thm:converge_sol} indicates that the truncated Poisson solution provides a uniformly accurate approximation to the true solution of the Poisson equation. Substituting this approximation into the control variate estimator \eqref{eq:modified_estimatorMH_stoch2} yields a modified estimator with controlled variance. We now investigate standard limit theory for additive functionals of Markov chains. Denote
$$
F_G := F(X) + \theta_1 \alpha(X,Y)(G(Y) - G(X)) 
+ \theta_2 (G(Y) - \E_{q(\cdot|X)}[G]).
$$
\begin{theorem}
    For any transition kernel $P$ and function $F$ under (A\ref{thm:A1}), (A\ref{thm:A2}) and (A\ref{thm:A3}), the estimator \eqref{eq:modified_estimatorMH_stoch2} follows, as $n \to \infty$,
    \begin{enumerate}

        \item Law of Large Numbers (LLN). $
        \mu_{n,G}(F) \xrightarrow{\mathrm{a.s.}} \pi(F).$
        \item Central Limit Theorem (CLT). $
        \sqrt{n}\big(\mu_{n,G}(F) - \pi(F)\big)
        \xrightarrow{D} \mathcal{N}(0, \sigma_{F_G}^2),$
        where $\sigma_{F_G}^2 <\infty$ denotes the asymptotic variance of $F_G$.
        \item Consistency of the coefficients. $(\theta_1,\theta_2)\xrightarrow{\mathrm{a.s.}}(\theta_1^*,\theta_2^*)$, where $(\theta_1^*,\theta_2^*)$ minimizes the asymptotic variance $\sigma_{F_G}^2$.
    \end{enumerate}
\end{theorem}


\subsection{Optimal scaling
\label{sec:optimal_scaling}
}

The parameter $\gamma$ plays the role of the global step size. Following optimal scaling results \citep{roberts2001optimal}, the standard practice for RWM and MALA is to tune $\gamma$ to reach acceptance rates $0.234$ and $0.574$. However, 
these guidelines may not be relevant
for the Gaussian invariant proposals, where optimal acceptance rates could vary with the target. An extreme case is obtained, by Proposition \ref{thm:gaussian_target}, when the target is Gaussian $\mathcal{N}(x|\mu,\Sigma)$, the acceptance rate is $1.0$ (since all samples are accepted) 
and the optimal step size is $\gamma=1$,
which results in i.i.d. sampling. 
For non-Gaussian targets, we observe in our experiments that by tuning  $\gamma$ to reach acceptance rates higher than those of RWM and MALA we achieve better performance for GI-RWM and GI-MALA. 
For example, in our GI-MALA experiments we found that the optimal $\gamma$ is obtained when the acceptance rate is around $0.8$. 
Consider another extreme case: Let the preconditioning matrix be the identity matrix $A_x = I_d$ in the GI-MALA proposal \eqref{eq:proposal_3}, and assume that the target distribution is non-Gaussian. In this case, as $d \rightarrow \infty$, the proposal \eqref{eq:proposal_3} converges to the standard MALA proposal. This is because the term involving $\gamma^2$ decays faster with dimension than the term involving $2\gamma$. Specifically, $\gamma^2$ scales as $O(d^{-2/3})$ while $2\gamma$ scales as $O(d^{-1/3})$.

In addition to the Gaussian target case and the infinite-dimensional case, a more interesting problem is to explore the behavior of optimal scaling with finite dimension and near-Gaussian target. Therefore, we investigate the theoretical optimal scaling on the factorized target
\begin{equation}\label{eq:near_gaussian_target}
    \pi_d(x) =\prod_{i=1}^d f(x_i)= \prod_{i=1}^d \exp\{g(x_i)\} = \prod_{i=1}^d\exp\{-\frac{x_i^2}{2}+\epsilon h(x_i) - \log(Z)\},
\end{equation}
where $Z$ is the normalizing constant, $h(x)$ is a polynomial with minimum order $4$ and $\epsilon\geq0$ is a small constant such that when $\epsilon=0$, the target becomes Gaussian. We assume that the density \eqref{eq:near_gaussian_target} is integrable and $g$ has the following properties:
\begin{assumption}[A4](\cite{roberts1998optimal})\label{thm:A4}
    $g$ is a $C^8$-function with derivatives $g^{(i)}$ satisfying
    $
        |g(x)|,|g^{(i)}(x)|\leq M_0(x),
    $
    for $1\leq i \leq 8$ and some integrable polynomial $M_0(x)$ in the target. We also assume that $g'$ is a Lipschitz function.
\end{assumption}
We use the inverse Fisher as the preconditioning matrix, which has diagonal entries 
\begin{equation}\label{eq:eniH_precondition}
    A_{ii} = E_{f}[-\nabla^2 g(x)]^{-1}=E_{f}[1-\epsilon h''(x)]^{-1}=1+\epsilon E_{f}[h'']+\epsilon^2 E_{f}[{h''}]^2 + O(\epsilon^3).
\end{equation}
\begin{theorem}\label{thm:optimal_scaling}
    For target \eqref{eq:near_gaussian_target} satisfying (A\ref{thm:A4}), proposal \eqref{eq:proposal_3} with preconditioning matrix \eqref{eq:eniH_precondition}, sufficiently large and small $d$ and $\epsilon$ respectively, the optimal $\gamma$ obtained from maximizing the global speed measure defined by \cite{titsias2019gradient}
    can be expressed by
    \begin{equation}\label{eq:objective}
        \gamma^* = \operatorname*{argmax}_{\gamma} ~ (2\gamma-\gamma^2)^{\kappa d/2}\Big(2\Phi(-\frac{\epsilon K \sqrt{d} \gamma^{3/2}}{2})-M\Big),
    \end{equation}
    where $\kappa$ is a hyperparameter, $M$ is a constant error bound for acceptance rate approximation, $\Phi(x)$ is CDF of standard normal distribution and
    $$K = \sqrt{\frac{5}{12}E_{f}[h'''(x)^2]+\frac{1}{4}Var_f[h''(x)]}>0.$$
\end{theorem}
The optimal value of $\gamma$ can be obtained using Algorithm 1 of \cite{titsias2019gradient}, a gradient-based adaptive MCMC method that updates $\gamma$ at each MCMC step using the speed measure gradient estimator. Moreover, we can also draw heuristic insights through numerical optimization under fixed settings. The value of the hyperparameter $\kappa$ serves as a trade-off between a high acceptance probability and a high entropy of the proposal distribution. Setting $\kappa = 2/d$, $K=2$ and $M = 0.001$, we numerically optimize the objective in \eqref{eq:objective} for various values of $\epsilon$ and $d$.
\begin{figure}[ht]
  \centering

  \begin{subfigure}[t]{0.49\textwidth}
    \centering
    \includegraphics[width=\linewidth]{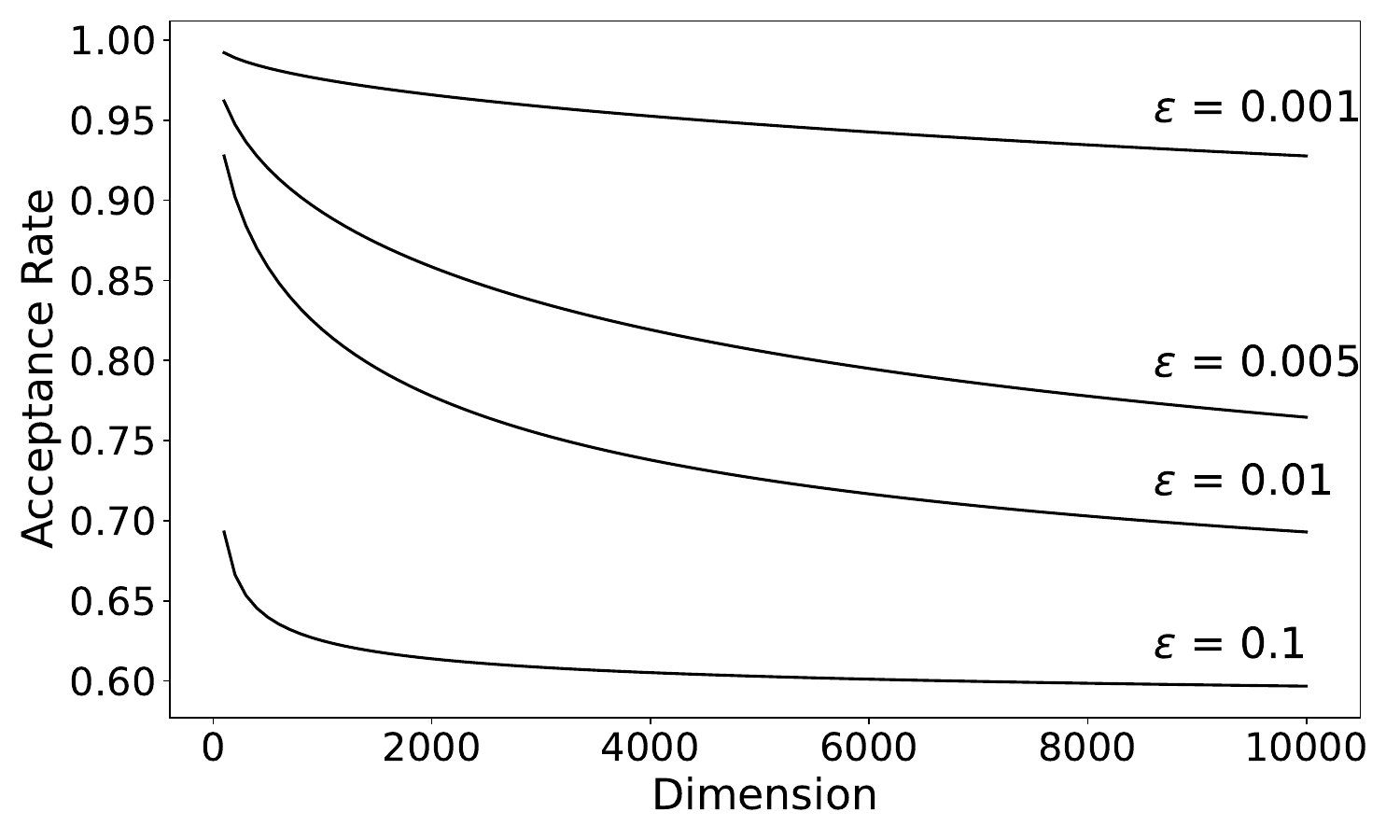}
    \caption{Acceptance rate}
  \end{subfigure}
  \hfill
  \begin{subfigure}[t]{0.49\textwidth}
    \centering
    \includegraphics[width=\linewidth]{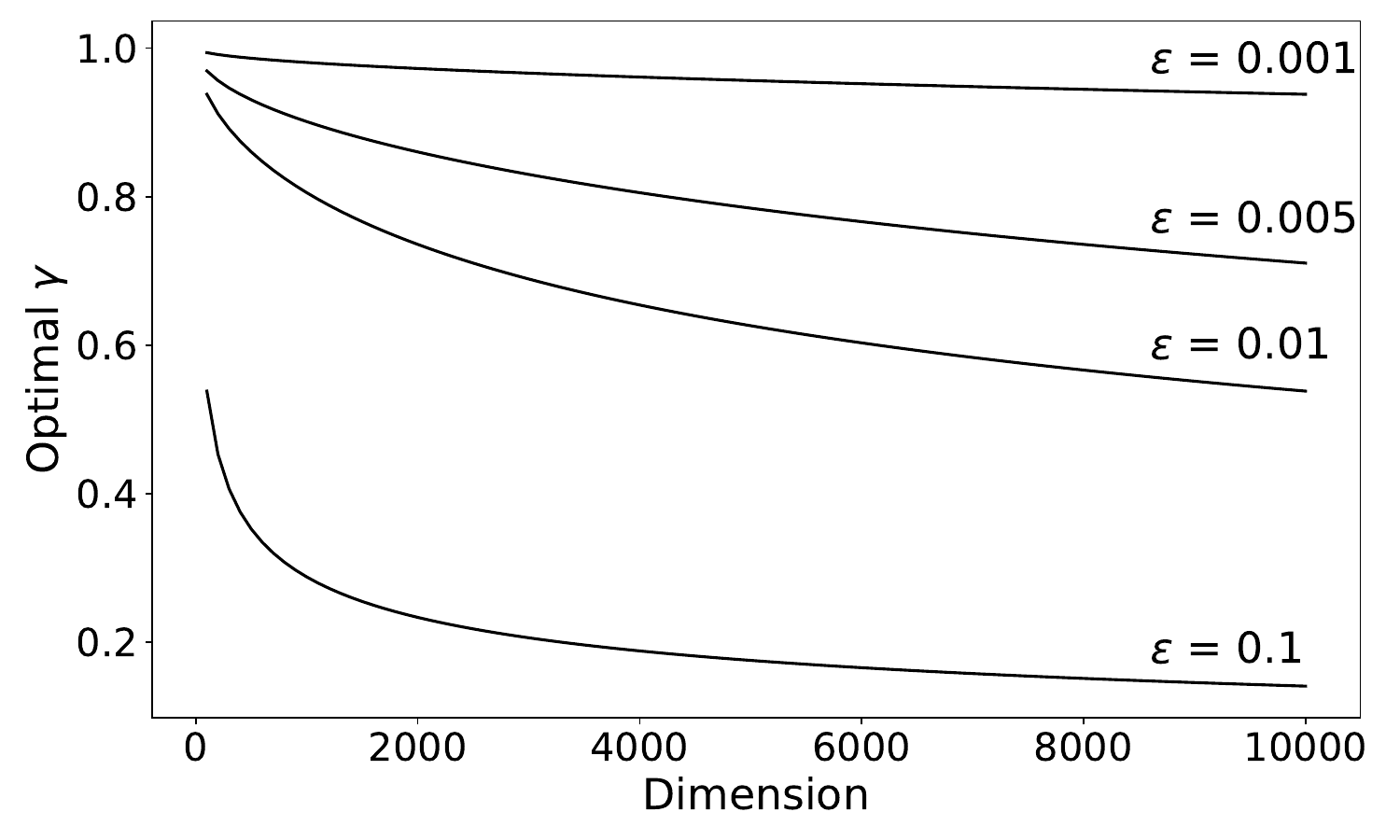}
    \caption{Optimal $\gamma$}
  \end{subfigure}
  \caption{The acceptance rates and optimal step sizes for various $\epsilon$ and dimensions.}\label{fig:optimal_scaling}
\end{figure}
Figure \ref{fig:optimal_scaling}(a) shows that when the target is close to Gaussian, we can adjust $\gamma$ such that the acceptance rate approaches $1$. However, as the dimension increases, the optimal acceptance rate decreases, and according to Figure \ref{fig:optimal_scaling}(b), $\gamma$ should be scaled accordingly with the dimension. As discussed previously, in the non-Gaussian case, GI-MALA gradually converges to MALA as the dimension increases.

As with similar studies, our analysis faces some limitations. First, the scope of the target distribution is restricted, as we only consider perturbations of a factorized normal distribution. Second, to reduce analytical complexity, we employ the expected preconditioning matrix, resulting in a position-independent formulation. Third, the definition of the speed measure does not precisely correspond to the speed of the limiting diffusion; consequently, the value of $\gamma$ obtained from \eqref{eq:objective} may not be truly optimal.

\section{Discussion
\label{sec:discussion}
}

We presented general-purpose Gaussian 
invariant MCMC algorithms, 
such as the gradient-based GI-MALA methods, and we studied similarities as well as differences with standard MALA samplers and second order or Hessian-based
methods.  An interesting property of ergodic estimation using these
samplers is that we can exploit
analytical solutions 
to the Poisson equation
to construct control variates
for variance reduction. 
Our main application in latent Gaussian 
models showed that we can  apply these methods to high-dimensional targets and achieve
large effective sample sizes and low cost variance reduction.  

In the theoretical results
the optimal scaling analysis reveals an important difference
between Gaussian invariant MCMC
and standard RWM and MALA schemes. While for RWM and MALA there exist well-established specific optimal acceptance rates based on diffusion limits 
\citep{roberts2001optimal}, for Gaussian invariant MCMC the acceptance rates depend on the Gaussianity of the target.  
Therefore, an approximate optimal scaling is derived under finite-dimensional assumptions, with a detailed characterization of its dependence on non-Gaussian features of the target density.

For future work it would be useful to extend the automatic selection of the position-dependent preconditioner $A_x$ in GI-MALA beyond latent Gaussian models, in order to balance computational cost and statistical efficiency for generic targets. Also it would be interesting to consider alternative ways to select the value of $\gamma$ by applying, for instance, some gradient-based optimization procedure to optimize the global speed measure.

\section*{Acknowledgments}
The second author acknowledges the support from the program “Drasi I” of the Athens University of Economics and Business.


\bibliography{references}

\appendix

\section{Further experiments}

\subsection{Binary GP classification}

Tables \ref{tab:german}, \ref{tab:pima} and \ref{tab:ripley} show the performance of the different sampling schemes in binary classification datasets with dimensions ranging from 250 to 1000. Table \ref{tab:app_GP_logistic_var} shows the estimated variance reduction factors.  

\begin{table}[H]
\caption{Comparison of sampling methods in German dataset. The size of the latent vector
$x$ is $d = 1000$. All numbers are averages across ten random repeats.}
\centering
\begin{tabular}{lllll}
\toprule
\em Method &\em Time(s)  &\em ESS (Min, Med, Max)  &\em Min ESS/s  \\ 
\midrule
GI-MALA  &  1.9  & (472.7, 1083.8, 1565.0)  &  244.12 (22.29)\\ 
MALA  &  2.8  & (79.3, 205.3, 333.7)  &  28.08 (5.81)\\ 
mGrad  &  2.0  & (177.5, 534.6, 938.2)  &  88.27 (9.14)\\ 
Ellipt  &  3.1  & (5.2, 19.9, 71.9)  &  1.67 (0.12)\\ 
pCN  &  1.0  & (4.4, 14.5, 56.8)  &  4.20 (0.17)\\ 
pCNL  &  2.6  & (6.9, 26.9, 87.4)  &  2.67 (0.25)\\ 
\bottomrule
\end{tabular}
\label{tab:german}
\end{table}

\begin{table}[H]
\caption{Comparison of sampling methods in Pima dataset. The size of the latent vector
$x$ is $d = 532$. All numbers are averages across ten random repeats.}
\centering
\begin{tabular}{lllll}
\toprule
\em Method &\em Time(s)  &\em ESS (Min, Med, Max)  &\em Min ESS/s  \\ 
\midrule
GI-MALA  &  1.0  &  (531.9, 1860.7, 2997.8)  &  519.76 (59.20)\\ 
MALA  &  1.4  &  (110.7, 259.8, 421.4)  &  79.81 (16.03)\\ 
mGrad  &  1.0  & (310.1, 971.1, 1672.8)  &  324.66 (30.36)\\ 
Ellipt  &  1.7  & (18.5, 66.9, 189.2)  &  10.92 (2.30)\\ 
pCN  &  0.6  &  (10.7, 42.8, 122.8)  &  18.46 (2.27)\\ 
pCNL  &  0.8  &  (24.9, 91.4, 226.4)  &  30.79 (6.43)\\ 
\bottomrule   
\end{tabular}
\label{tab:pima}
\end{table}

\begin{table}[H]
\caption{Comparison of sampling methods in Ripley dataset. The size of the latent vector
$x$ is $d = 250$. All numbers are averages across ten random repeats.}
\centering
\begin{tabular}{lllll}
\toprule
\em Method &\em Time(s)  &\em ESS (Min, Med, Max)  &\em Min ESS/s  \\ 
\midrule 
GI-MALA  &  0.3  & (50.4, 314.1, 1999.0)  &  151.51 (62.24)\\ 
MALA  &  0.5  &  (31.3, 170.9, 770.4)  &  68.28 (17.71)\\ 
mGrad  &  0.3  &  (52.0, 294.4, 1732.8)  &  149.68 (41.68)\\ 
Ellipt  &  0.5  & (11.2, 50.5, 210.8)  &  21.45 (2.73)\\ 
pCN  &  0.2  &  (7.3, 32.0, 130.9)  &  36.42 (6.93)\\ 
pCNL  &  0.3  &  (12.8, 62.9, 274.2)  &  44.41 (10.20)\\ 
\bottomrule
\end{tabular}
\label{tab:ripley}
\end{table}

\begin{table}[H]
\centering
\caption{Range of estimated variance reduction factors for Gaussian process binary classification targets.
}
\begin{tabular}{llllll}
\toprule
  Dataset  & $n=1,000$& $n=10,000$ & $n=50,000$ & $n=200,000$     &  \\
\midrule
Ripley ($d=250$) & 1.11-3.54 & 1.14-3.15 & 1.13-3.38 & 1.13-3.14 \\
Pima ($d=532$) & 1.93-5.92 & 2.04-5.88 & 1.79-5.37 & 1.87-6.06 \\
German ($d=1000$) & 2.28-5.91 & 2.15-5.83 & 2.24-5.72 & 2.24-6.68 \\
\bottomrule
\end{tabular}
\label{tab:app_GP_logistic_var}
\end{table}

\subsection{Variance reduction based on the truncated solution}

Figure 1 in Section 5.2.1 visualizes the variance ratios across $\nu$ for $N=2$ and $N=5$. To make the comparison more transparent, we also provide Table~\ref{tab:var_ratio_t}, which lists the corresponding values for all $b\in\{0,1,2,3\}$.

\begin{table}[H]
\centering
\caption{Estimated factors (variance ratios) by which the variance of $\mu_{n}(F)$ is larger than the variance of $\mu_{n,G}(F)$ when the target is a univariate Student's t-distribution with $\nu$ degrees of freedom and $F(x)= I(x > b)$, for different values of $\nu$, $b$ and $N$.
\label{tab:var_ratio_t}
}
\centering

\begin{tabular}{lcccc}
\toprule
$\nu$ & $c=0$ & $c=1$ & $c=2$ & $c=3$ \\
\midrule
1 & 1.05 & 1.02 & 1.03 & 1.01 \\
2 & 1.21 & 1.26 & 1.02 & 1.05 \\
5 & 2.61 & 1.89 & 1.49 & 1.19 \\
30 & 61.76 & 24.56 & 7.14 & 2.45 \\
100 & 964.95 & 287.37 & 74.59 & 12.10 \\
300 & 4790.72 & 1992.52 & 540.42 & 62.21 \\
1000 & 27546.21 & 14937.28 & 5557.87 & 839.83 \\
\bottomrule
\end{tabular}
\begin{tabular}{lcccc}
\toprule
$\nu$ & $c=0$ & $c=1$ & $c=2$ & $c=3$ \\
\midrule
1 & 1.03 & 1.01 & 1.03 & 1.01 \\
2 & 1.17 & 1.13 & 1.11 & 1.08 \\
5 & 2.90 & 2.17 & 1.40 & 1.30 \\
30 & 70.63 & 28.37 & 6.94 & 2.45 \\
100 & 710.52 & 259.85 & 65.28 & 8.24 \\
300 & 7821.55 & 2873.60 & 551.41 & 103.86 \\
1000 & 76857.98 & 38228.00 & 4109.02 & 889.97 \\
\bottomrule
\end{tabular}
\end{table}

\begin{figure}[H]
    \centering
    \includegraphics[width=0.75\linewidth]{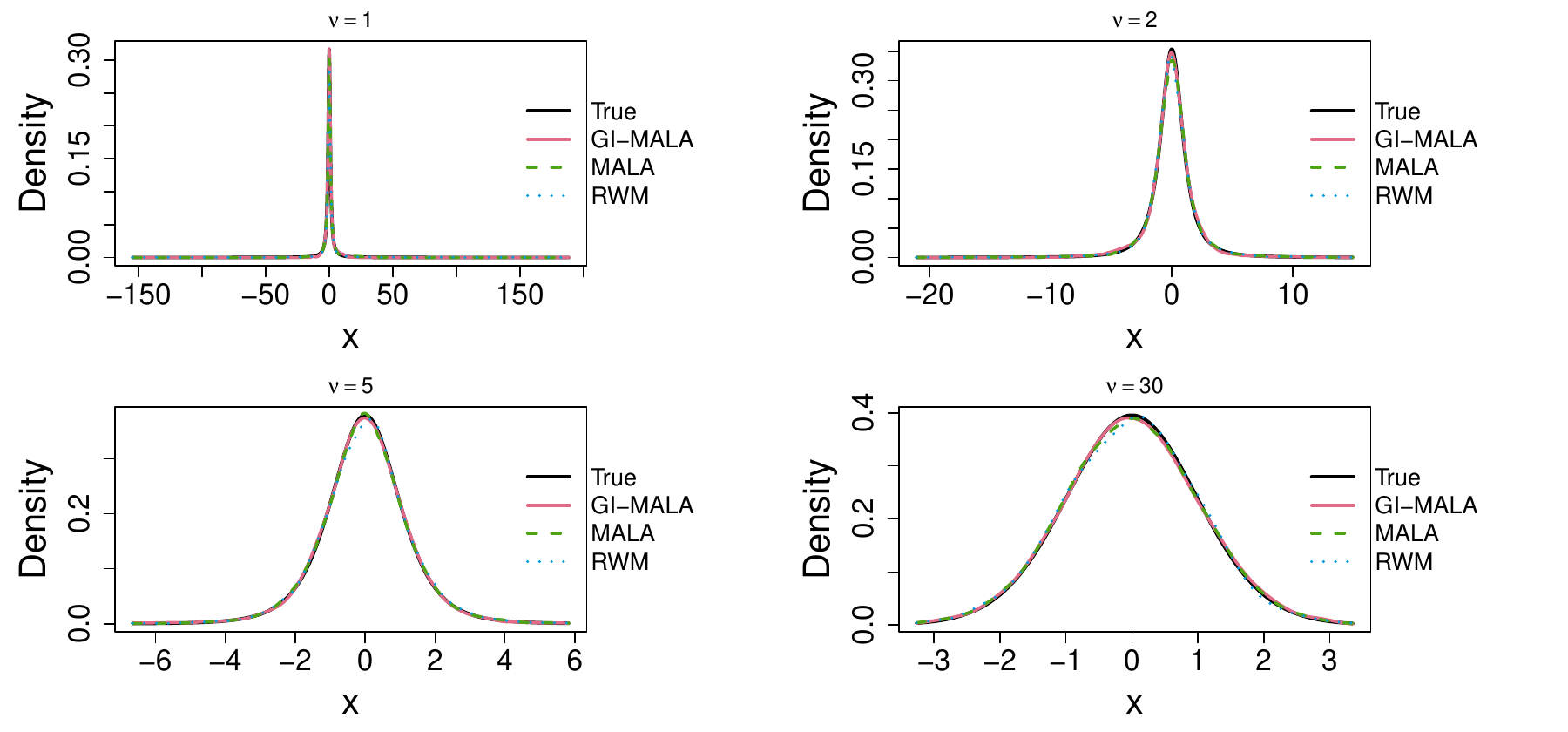}
    \caption{Estimated stationary densities (kernel density estimates) from \textsc{GI-MALA}, \textsc{MALA}, and \textsc{RWM} targeting univariate Student's $t$-distributions, overlaid with the true target density.}
    \label{fig:tab:ess_t_selftuned_gi_0p95}
\end{figure}

We also compare \textsc{GI-MALA}, preconditioned \textsc{MALA} and \textsc{RWM} in terms of convergence to the target and sampling efficiency for univariate Student's $t$-distributions. Figure~\ref{fig:tab:ess_t_selftuned_gi_0p95} shows that the empirical stationary densities produced by \textsc{GI-MALA}, \textsc{MALA}, and \textsc{RWM} closely match the true $t_\nu$ target across all $\nu$, indicating that all methods converge to the correct distribution.

Table~\ref{tab:ess_t_selftuned_gi_0p95} highlights a clear efficiency trade-off with tail-heaviness: for heavy-tailed targets (small $\nu$, e.g.\ $\nu=1,2$) \textsc{RWM} attains the largest ESS/s, whereas as the target becomes closer to Gaussian (large $\nu$, e.g.\ $\nu=30,100,1000$) \textsc{GI-MALA} becomes the most efficient, achieving the highest ESS/s among the three samplers.

\begin{table}[H]
\centering
\caption{ESS, acceptance rate, and step size $\gamma$ for samplers targeting a univariate Student's $t$-distribution. Each method adapts $\gamma$ during burn-in (3000 iterations) and then freezes it. Results are averages over 10 independent runs of total length 7000.}
\label{tab:ess_t_selftuned_gi_0p95}
\begin{tabular}{llccc}
\toprule
 &  & GI-MALA & MALA & RWM \\
\midrule
$\nu=1$ & ESS      & 60.80 & 177.20 & 303.38 \\
 & Acc(post)& 0.978 & 0.602 & 0.283 \\
 & $\gamma$ & 0.3268 & 5.8748 & 47.5163 \\
$\nu=2$ & ESS      & 160.43 & 749.31 & 862.53 \\
 & Acc(post)& 0.979 & 0.594 & 0.280 \\
 & $\gamma$ & 0.3175 & 3.7958 & 19.4026 \\
$\nu=5$ & ESS      & 690.76 & 3685.96 & 1808.54 \\
 & Acc(post)& 0.978 & 0.604 & 0.278 \\
 & $\gamma$ & 0.3671 & 2.4538 & 12.2515 \\
$\nu=30$ & ESS      & 4648.17 & 7275.93 & 1802.58 \\
 & Acc(post)& 0.980 & 0.602 & 0.277 \\
 & $\gamma$ & 0.8264 & 1.7474 & 9.4242 \\
$\nu=100$ & ESS      & 10000.00 & 7007.63 & 1841.90 \\
 & Acc(post)& 0.982 & 0.599 & 0.273 \\
 & $\gamma$ & 1.5626 & 1.6710 & 9.7925 \\
$\nu=1000$ & ESS      & 10000.00 & 7046.73 & 1731.22 \\
 & Acc(post)& 0.984 & 0.602 & 0.278 \\
 & $\gamma$ & 1.9995 & 1.6108 & 9.3024 \\
\bottomrule
\end{tabular}
\end{table}

\subsection{Empirical convergence analysis}

\begin{figure}[H]
\centering
\begin{subfigure}[b]{0.45\linewidth}
  \centering
  \includegraphics[width=\linewidth]{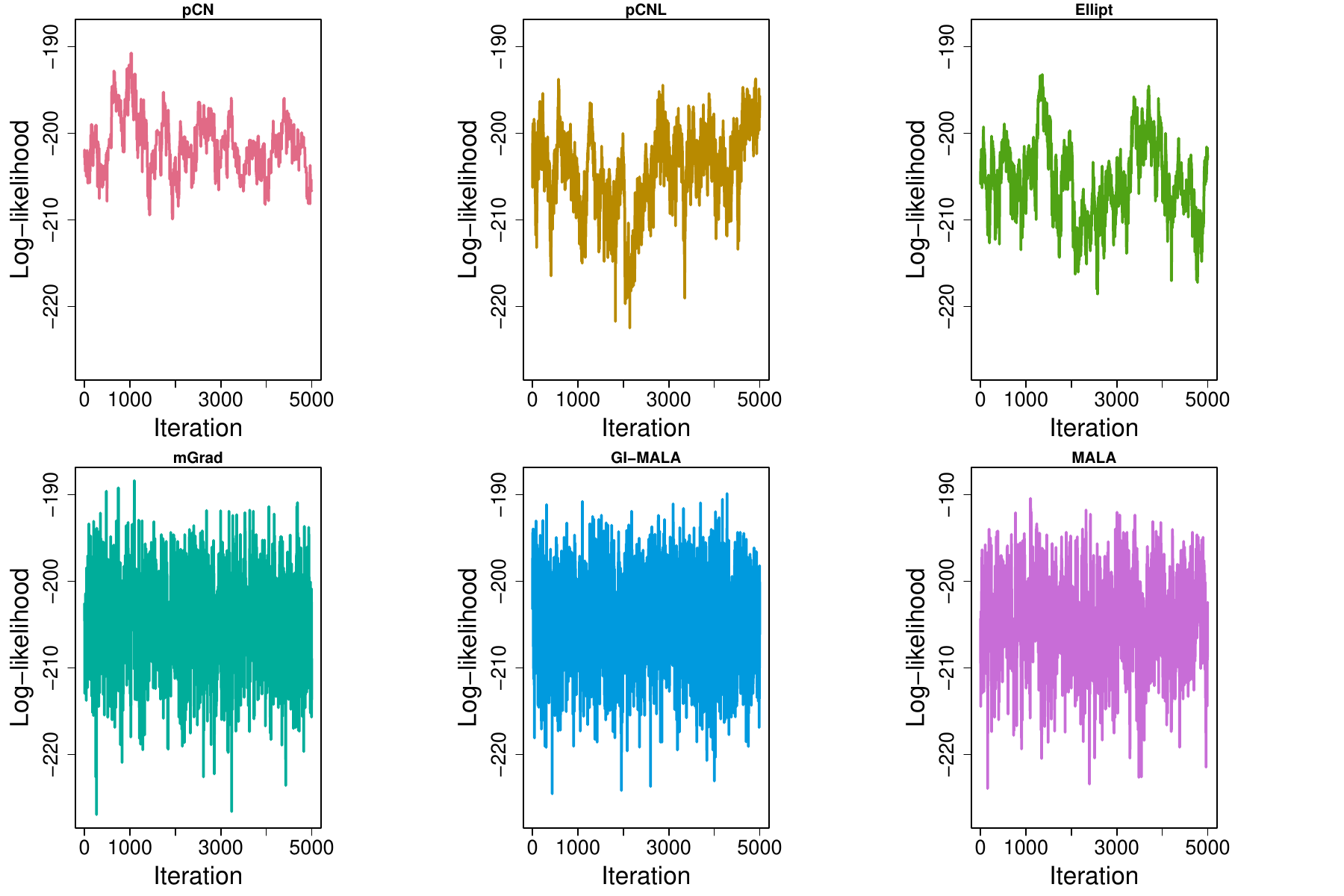}
  \caption{Trace plots of the log-likelihood from a single run of the different algorithms.}
\end{subfigure}\hfill
\begin{subfigure}[b]{0.45\linewidth}
  \centering
  \includegraphics[width=\linewidth]{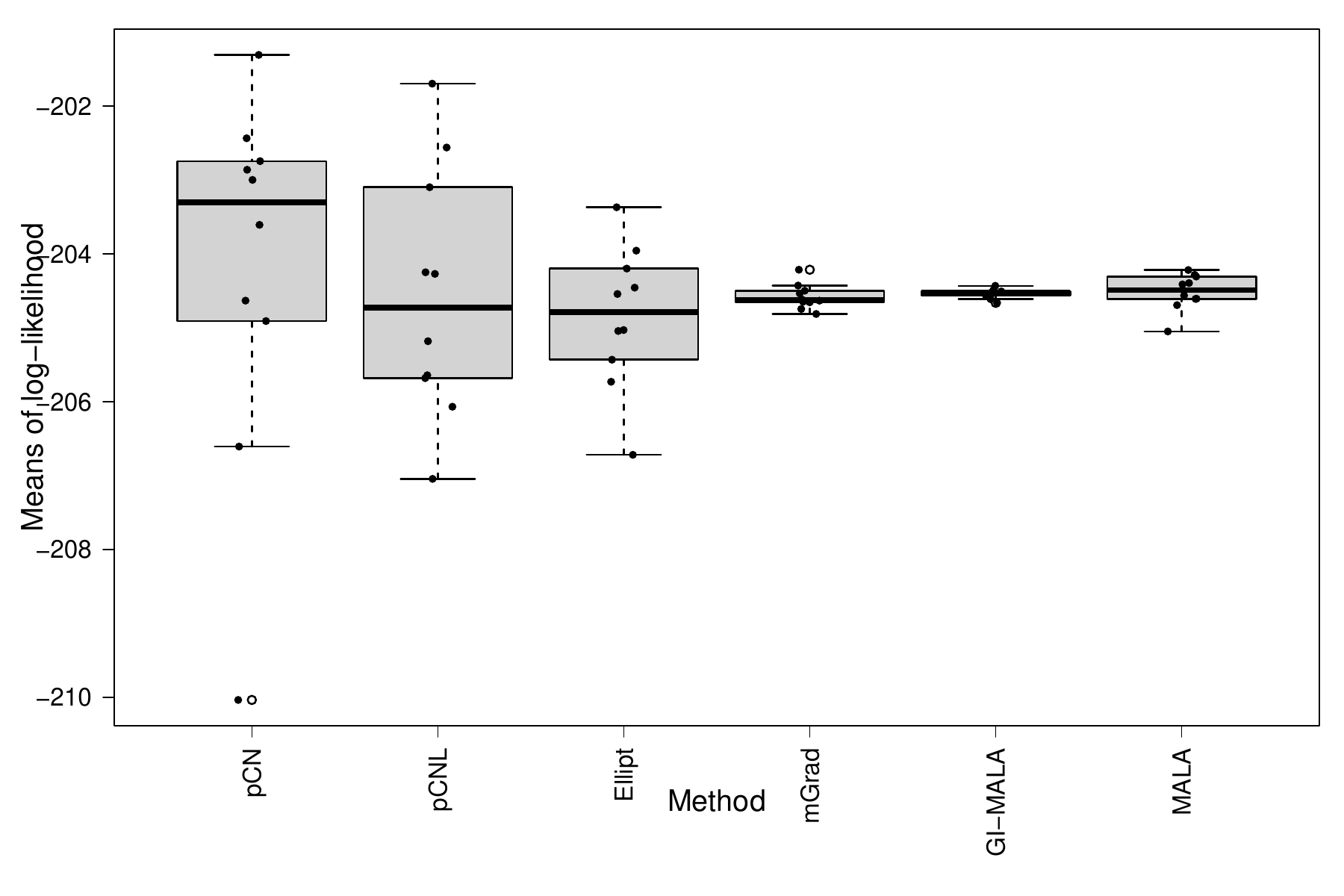}
  \caption{The means of the log-likelihood across 10 repetitions of the different algorithms.}
\end{subfigure}
\caption{Australian dataset.}
\label{fig:diag-australian}
\end{figure}

\begin{figure}[H]
\centering
\begin{subfigure}[b]{0.45\linewidth}
  \centering
  \includegraphics[width=\linewidth]{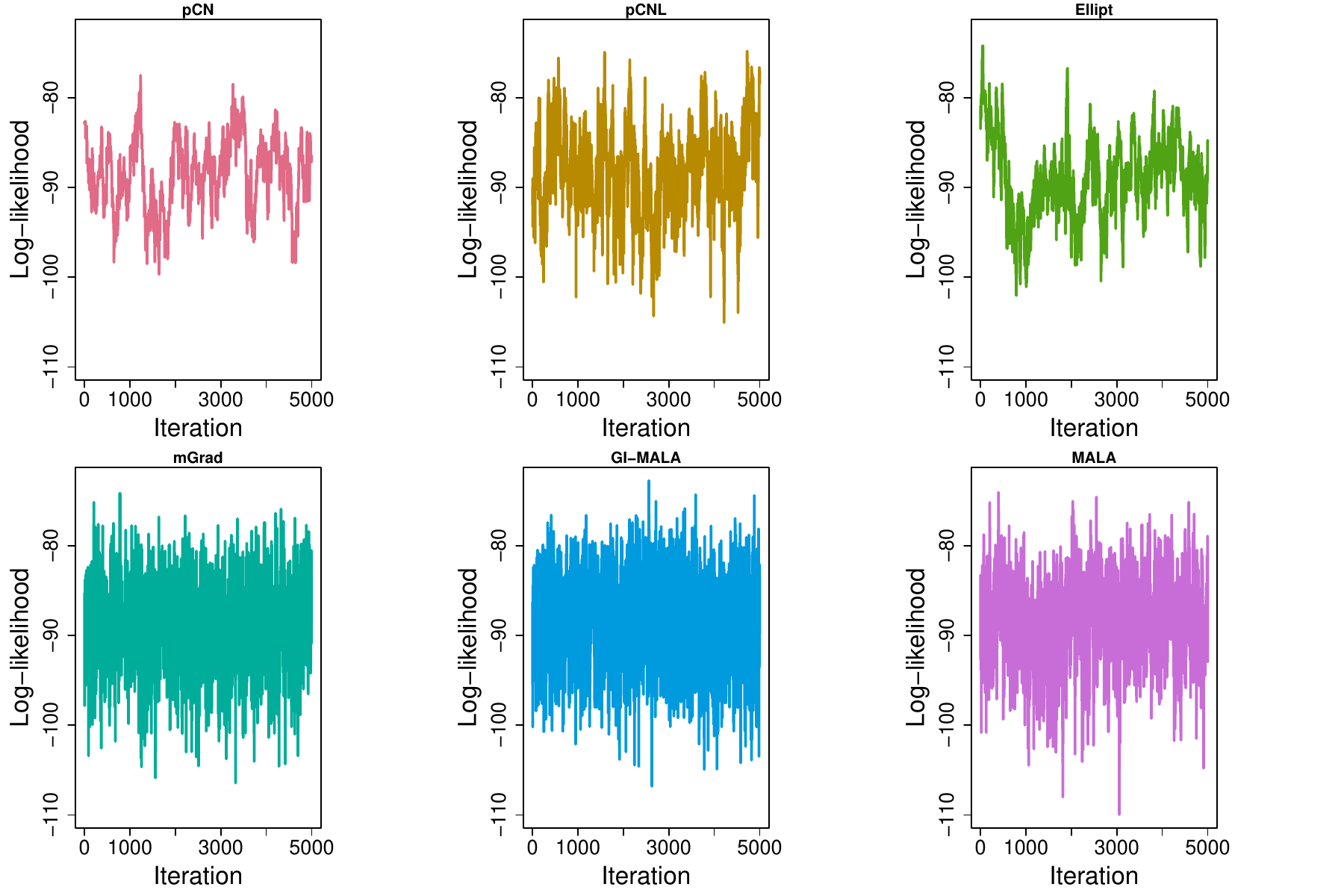}
  \caption{Trace plots of the log-likelihood from a single run of the different algorithms.}
\end{subfigure}\hfill
\begin{subfigure}[b]{0.45\linewidth}
  \centering
  \includegraphics[width=\linewidth]{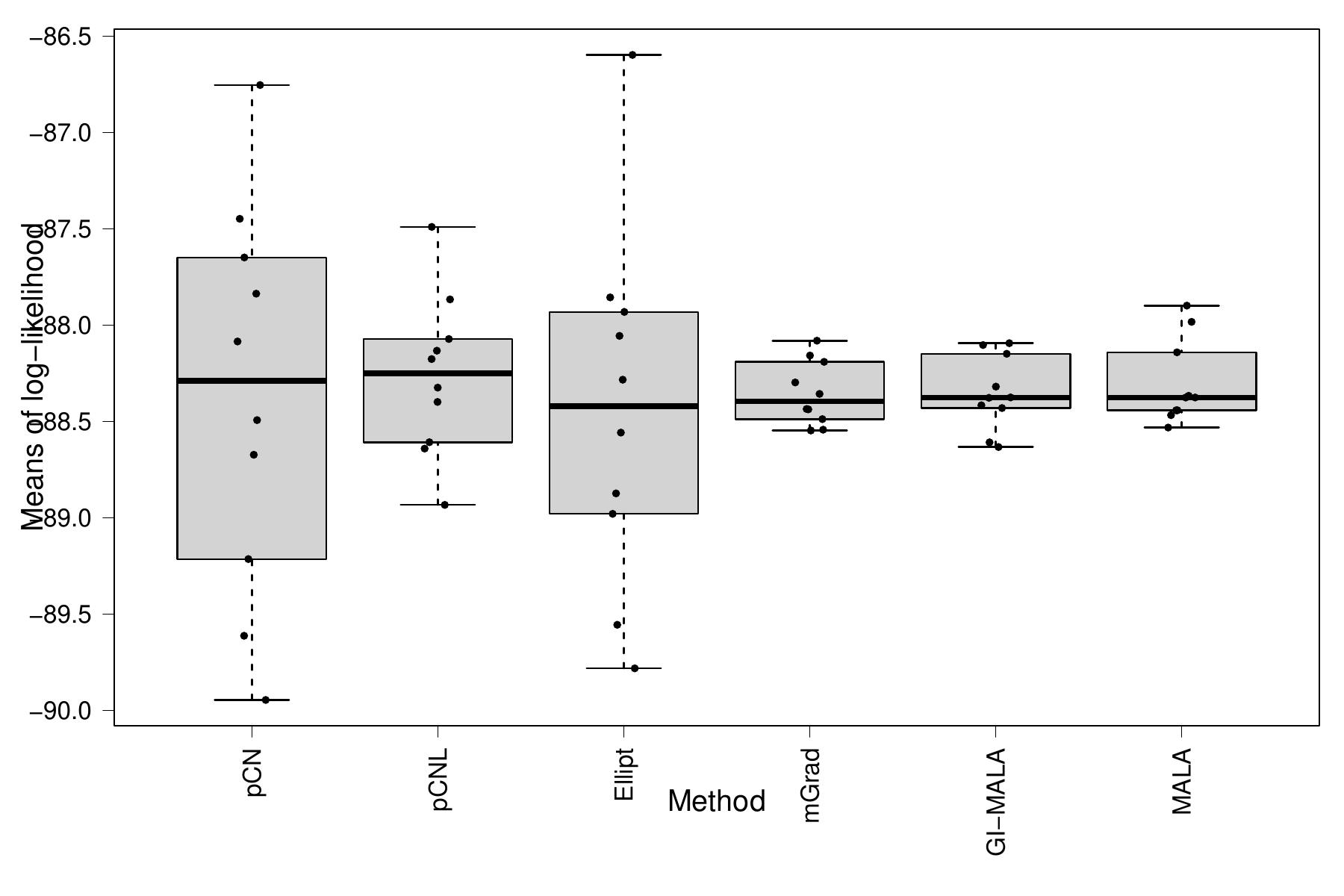}
  \caption{The means of the log-likelihood across 10 repetitions of the different algorithms.}
\end{subfigure}
\caption{Heart dataset.}
\label{fig:diag-heart}
\end{figure}

\begin{figure}[H]
\centering
\begin{subfigure}[b]{0.45\linewidth}
  \centering
  \includegraphics[width=\linewidth]{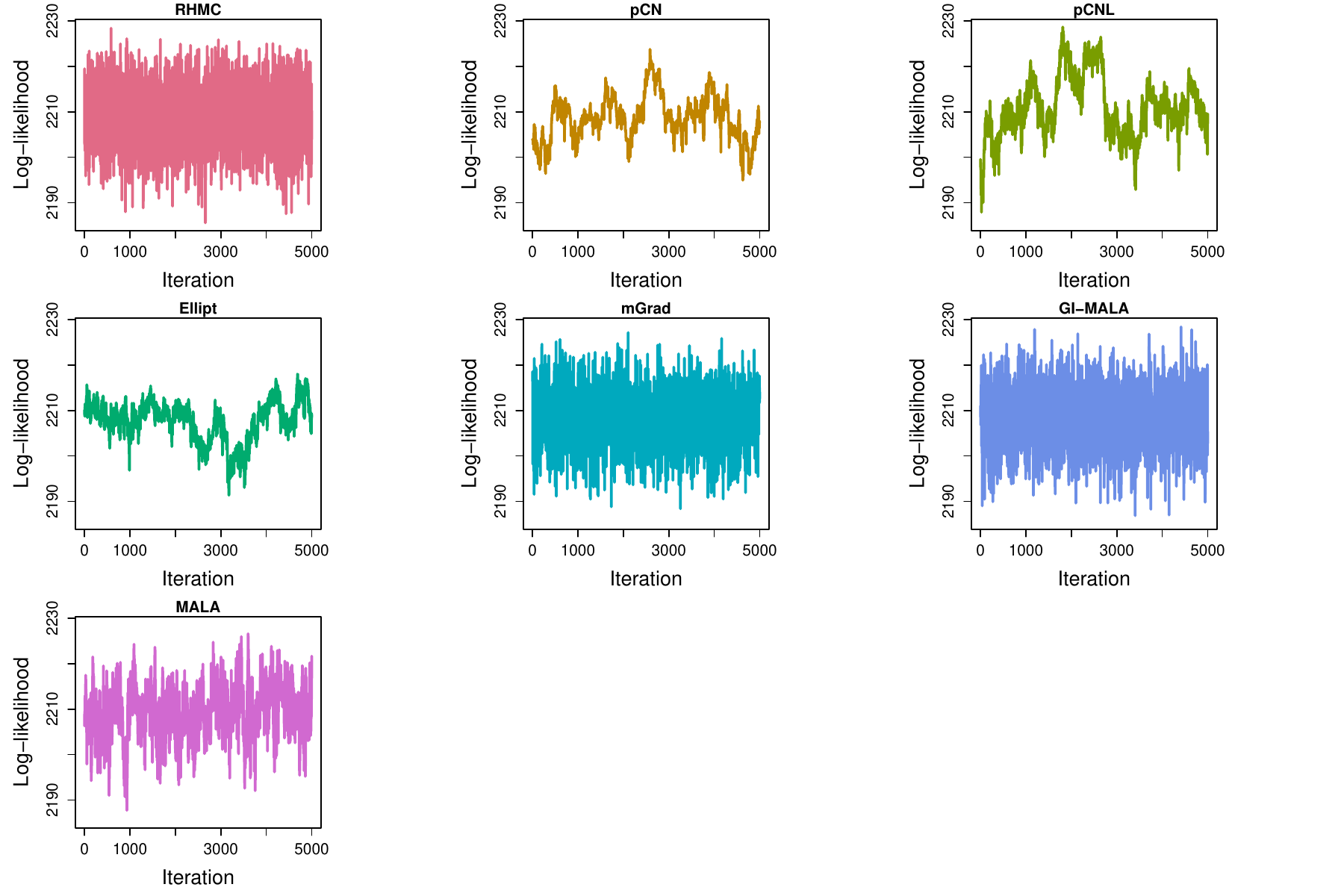}
  \caption{Trace plots of the log-likelihood from a single run of the different algorithms.}
\end{subfigure}\hfill
\begin{subfigure}[b]{0.45\linewidth}
  \centering
  \includegraphics[width=\linewidth]{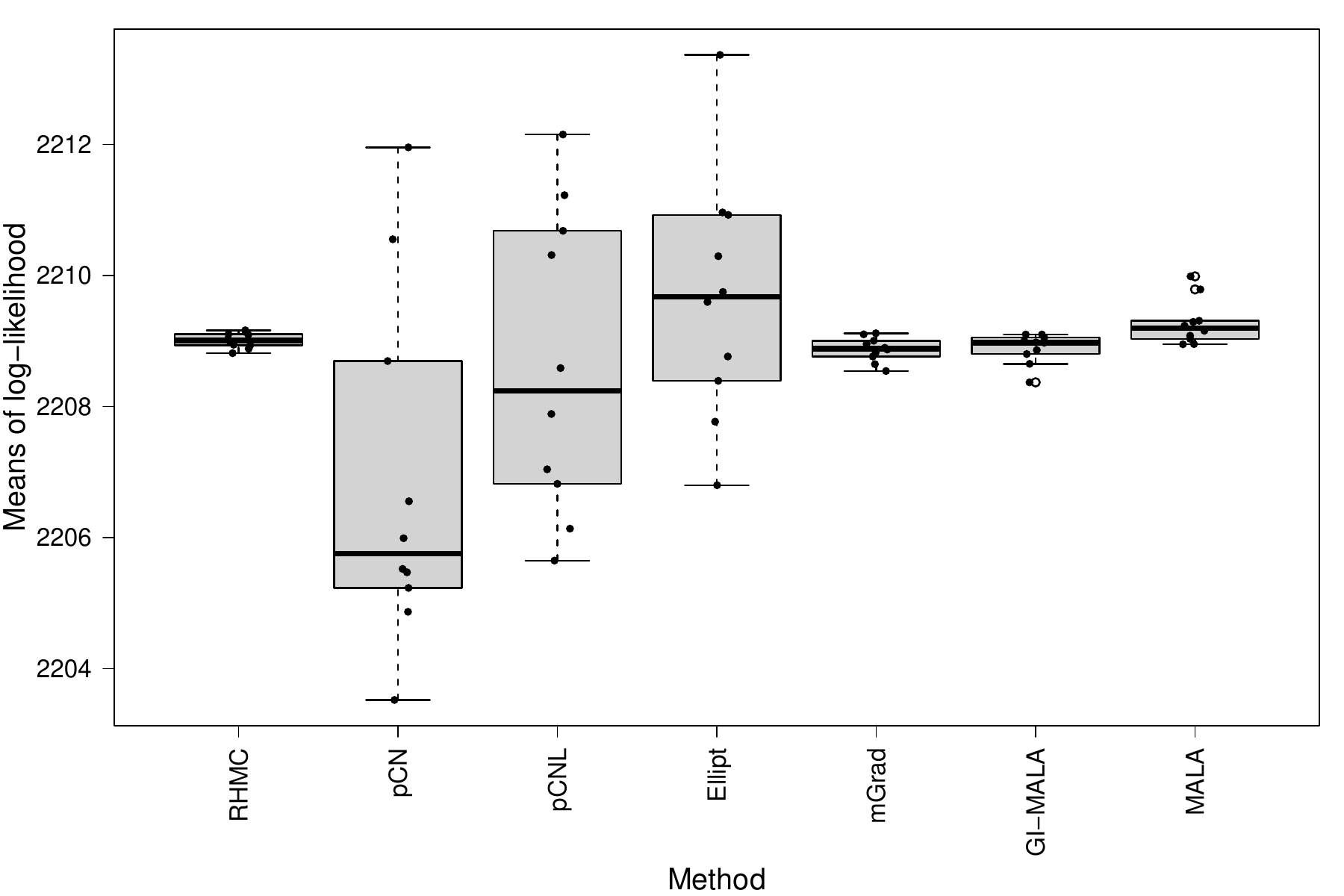}
  \caption{The means of the log-likelihood across 10 repetitions of the different algorithms.}
\end{subfigure}
\caption{Log-Gaussian Cox process dataset.}
\label{fig:diag-cox}
\end{figure}

To assess agreement across algorithms and the variability of the resulting estimates, we rely on the 10 replicates of each experiment configuration and compare the log-target ($\log \pi(x)$) behaviour: (i) within a single run and (ii) across the 10 independent repetitions. For the three datasets (Australian, Heart, and the log-Gaussian Cox process), the single-run trace plots show that the log-target values stabilizes in a comparable range across methods (Figs.~\ref{fig:diag-australian}a, \ref{fig:diag-heart}a and \ref{fig:diag-cox}a), which serves as a practical consistency check that all samplers converge
to the target distribution.  Across the 10 repeats, the boxplots of the mean log-target (averaged over 5000 iterations) show noticeably larger dispersion for pCN/pCNL/Ellipt, whereas the gradient-based samplers (mGrad, GI-MALA and RHMC) exhibit much tighter clustering (Figs.~\ref{fig:diag-australian}b, \ref{fig:diag-heart}b and \ref{fig:diag-cox}b), indicating better Monte Carlo estimates. These boxplots correlate well with 
the trace log-target plots where pCN/pCNL/Ellipt clearly have higher autocorrelation.

\subsection{Some additional experiments}

\subsubsection{Gaussian process regression}
\label{sec:gp_regression}

Following \cite{titsias2018auxiliary} we compare the competitive samplers under different levels of the likelihood informativeness. We consider a Gaussian process regression setting where $g(x)$  
  has the form
\begin{equation}
g(x) = - \frac{d}{2} \log(2 \pi \sigma^2) - \frac{1}{2 \sigma^2} \sum_{i=1}^d (y_i - x_i)^2  
\label{eq:fxreg}
\end{equation}    
and $y_i$ is a real-valued observation. We assume that $x \sim \mathcal{N}(0,\Sigma_0)$ with $\Sigma_0$ defined via the squared exponential kernel function $c(z_i,z_j) = \sigma_x^2 \exp\{ - \frac{1}{2 \ell^2} ||z_i  - z_j||^2 \}$, $z_i$ is a input vector associated with the latent function value $x_i$. Notice that in the specified model $\sigma^2$ is depicts the level of information carried on the likelihood for the observations. Table \ref{tab:GaussReg} presents the ESS, running times and
an overall efficiency score for the very challenging example with $\sigma^2=0.01$ and $d=1000$. 
It is clear that the highest ESS is achieved by the proposed GI-MALA while the mGrad \citep{titsias2018auxiliary} and MALA which only differs in the scaling factor with GI-MALA have similar ESS which is about $4$ times smaller than the ESS of the proposed sampler. Importantly, these results are not different even if we take into account computational time; see the last column in Table \ref{tab:GaussReg} where the ratio between minimum ESS (among all dimensions of $x$) over running time is presented; throughout the rest of the paper we refer to this ratio as overall efficiency score of a method.

\begin{table}[H]
\caption{Comparison of sampling methods in the regression dataset with $\sigma^2 =0.01$ and $d=1000$. All numbers are averages across ten repeats.}
\centering
\begin{tabular}{lllll}
\toprule
\em Method &\em Time(s)   &\em ESS (Min, Med, Max)  &\em Min ESS/s  \\ 
\midrule
GI-MALA  &  5.4 & (4369.5, 4906.7, 5000.0)  &  832.34 (128.38)\\ 
MALA  &  14.4  & (145.7, 218.2, 279.7)  &  10.11 (1.08)\\ 
mGrad  &  9.3  & (835.3, 1105.8, 1301.7)  &  90.94 (12.06)\\ 
Ellipt  &  2.6  & (8.9, 36.2, 134.5)  &  3.53 (1.10)\\ 
pCN  &  2.1  & (5.7, 32.6, 124.5)  &  2.92 (1.14)\\ 
pCNL  &  14.9  & (10.3, 53.5, 192.6)  &  0.69 (0.15)\\ 
\bottomrule
\end{tabular}
\label{tab:GaussReg}
\end{table}

\section{Proofs}

\subsection{Proof of Proposition 1 
\label{app:proposition2}}
The main claim is that the $n$-step transition  
is $
P^n(x_n | x) = \mathcal{N}(x_n| \beta^n x + (1 - \beta^n) \mu, (1 - \beta^{2 n}) \Sigma)
$. First note that
due to the Gaussian invariance property all samples are accepted. Therefore, the MCMC transition density coincides with the Gaussian invariant proposal. Given the initial state $x$, the next MCMC state $x_1$ follows
$$
P(x_1 | x) = \mathcal{N}(x_1| \beta x + (1 - \beta) \mu, (1 - \beta^{2}) \Sigma)
$$
and subsequently $x_2$ follows 
$$
P(x_2 | x) = \int
\mathcal{N}(x_2| \beta x_1 + (1 - \beta) \mu, (1 - \beta^{2}) \Sigma)
\mathcal{N}(x_1| \beta x + (1 - \beta) \mu, (1 - \beta^{2}) \Sigma) d x_1.
$$
Thus, $x_2$ follows a Gaussian with mean $\E[x_2|x] = \beta \E[x_1|x] + (1- \beta) \mu =\beta \left(
\beta x + (1-\beta) \mu\right) + (1-\beta) \mu = \beta^2 x + (1 - \beta^2) \mu$. To simplify the covariance
$\E\left[ (x_2 - \E[x_2|x]) 
(x_2 - \E[x_2|x])^\top
\right]$ we substitute  
$x_2 = \beta x_1 + (1-\beta) \mu + \epsilon_1$ and  $\E[x_2|x] = \beta \E[x_1|x] + (1- \beta) \mu$ where $\epsilon_1 \sim \mathcal{N}(0, (1-\beta^2)\Sigma)$, and 
obtain $\E\left[ (\beta x_1 + \epsilon_1 - \beta \E[x_1|x]) 
(\beta x_1 + \epsilon_1 - \beta \E[x_1|x])^\top
\right] =  \beta^2 \E\left[ (x_1  -  \E[x_1|x]) 
(x_1 - \E[x_1|x])^\top
\right] + \E\left[\epsilon_1 \epsilon_1^\top
\right] =
\beta^2 (1-\beta^2)\Sigma + (1-\beta^2) \Sigma = (1- \beta^4) \Sigma$. Following 
the same logic we can recursively prove the formula for the $n$-step transition. 
The second part of the  Proposition 1 is trivial 
since it just substitutes the Gaussian invariant transitions to the Poisson solution.

\subsection{Proof of Proposition 2 and MH ratio of MALA in latent Gaussian models 
\label{app:GP_prop_properties}
} 

We first prove (i). Recall the form of the proposal 
$$
 q(y | x) = \dN(y | x + \gamma A_x(\nabla g(x) - \Sigma_0^{-1} x), (2 \gamma - \gamma^2) A_x ).
$$
The mean can be written as 
$x + \gamma A_x(\nabla g(x) - \Sigma_0^{-1} x - \delta_x x + \delta_x x) = x + 
\gamma A_x(\nabla g(x) - A_x^{-1} x + \delta_x x)$ 
which further simplifies as 
$(1- \gamma) x + \delta_x \gamma A_x (x + \delta_x^{-1} \nabla g(x))$. Since  $\Sigma_0 = U \Lambda U^\top$
we have
$$
A_x = (\Sigma_0^{-1} + \delta_x)^{-1} = \frac{1}{\delta_x} U \Lambda (\Lambda + \delta_x^{-1} I)^{-1}  U^\top.   
$$
Then, if we denote  
 $\zeta_x = U^\top (x + \delta_x^{-1} \nabla g(x))$,
 we observe that the mean of the proposal is written as
 $
 (1-\gamma) x  
+  U  \gamma \Lambda (\Lambda + \delta_x^{-1} I  )^{-1} \zeta_x,   
$
while 
$$
y = (1-\gamma) x  
+  U  \left[ \gamma \Lambda (\Lambda + \delta_x^{-1} I  )^{-1} \zeta_x + \sqrt{\delta_x^{-1} (2 \gamma - \gamma^2)} \Lambda^{\frac{1}{2}} (\Lambda  + \delta_x^{-1} I )^{-\frac{1}{2}} 
\epsilon \right],   
\ \ \epsilon \sim \mathcal{N}(0,I_d)
$$
generates a sample from the proposal. Note that the computation of $y$ requires 
only a single matrix vector multiplications of cost $O(d^2)$ to obtain $U \times [\ldots]$. The rest operations are $O(d)$ since $\Lambda$ is diagonal matrix. 
Note also that $\zeta_x$ has been precomputed and stored from the previous step; see next.  

To prove (ii) we need to  show that the ratio 
$\frac{ N(y|0, \Sigma_0)  q(x|y)}{N(x | 0, \Sigma_0) q(y|x)} 
= \exp\{h(x,y) - h(y,x)\}$.
We first observe that an alternative way to write the 
proposal is as
$$
q(y|x) = \frac{\mathcal{N}\left(y|x  + (\gamma / \delta_x) \nabla g(x), \frac{2 \gamma - \gamma^2}{\delta_x}\right) \mathcal{N}(y| (1-\gamma) x, (2\gamma -\gamma^2) \Sigma_0)}{\mathcal{N}\left(0|
\gamma \zeta_x, (2 \gamma - \gamma^2) (\Sigma_0 + \delta_x^{-1})   
\right)}.
$$
Since $\mathcal{N}(y| (1-\gamma) x, (2\gamma -\gamma^2) \Sigma_0)$ is invariant to the 
Gaussian prior, these terms 
together with the Gaussian 
priors cancel out in the ratio which simplifies as  
$$
\frac{ N(y|0, \Sigma_0)  q(x|y)}{N(x | 0, \Sigma_0) q(y|x)}  
= 
\frac{\mathcal{N}(x|y  + (\gamma / \delta_y) \nabla g(y), \frac{2 \gamma - \gamma^2}{\delta_y})} {\mathcal{N}(0|
\gamma \zeta_y, (2 \gamma - \gamma^2) (\Sigma_0 + \delta_y^{-1}) )}
\times 
\frac{\mathcal{N}(0|
\gamma \zeta_x, (2 \gamma - \gamma^2) (\Sigma_0 + \delta_x^{-1}))} 
{\mathcal{N}(y|x  + (\gamma / \delta_x) \nabla g(x), \frac{2 \gamma - \gamma^2}{\delta_x})}
$$
from which is straightforward to obtain the final result
$\frac{ N(y|0, \Sigma_0)  q(x|y)}{N(x | 0, \Sigma_0) q(y|x)} 
= \exp\{h(x,y) - h(y,x)\}$
where
$$
h(x,y) 
 = \frac{1}{2} \frac{\delta_x}{(2\gamma - \gamma^2)} ||y - x - \frac{\gamma}{\delta_x} \nabla g(x) ||^2 -\frac{1}{2} \sum_{i=1}^d \log (\lambda_i \delta_x + 1)  - \frac{1}{2}\frac{\gamma}{(2 - \gamma)} \zeta_x^\top \left(\Lambda + \delta_x^{-1} I_d \right)^{-1} \zeta_x.
$$
To obtain the ratio we need to perform a single $O(d^2)$ matrix 
vector multiplication to obtain 
$\zeta_y$ for the newly proposed 
sample, while all remaining computations are $O(d)$. 
Therefore, in total only two 
$O(d^2)$ operations are needed per MCMC iteration.

Next, we also compute the MH ratio in the case of the preconditioned MALA algorithm where the proposal distribution is
\begin{equation*}\label{eq:GP_prop_MALA}
q(y \mid x) \;=\; \mathcal N\!\bigl(y \mid x + \gamma A_x(\nabla g(x) - \Sigma_0^{-1}x),\, 2\gamma A_x\bigr).
\end{equation*}

Let $\kappa = 2\gamma$. Then
\begin{equation*}\label{eq:q-decomp}
\begin{aligned}
q(y\mid x)
&=
\frac{
\mathcal N\!\left(y\mid x+\frac{\gamma}{\delta_x}\nabla g(x),\ \frac{\kappa}{\delta_x}I\right)\;
\mathcal N\!\left(y\mid (1-\gamma)x,\ \kappa\Sigma_0\right)}
{\mathcal N\!\left(0\mid \gamma\zeta_x,\ \kappa(\Sigma_0+\delta_x^{-1}I)\right)},\,\,\,
\zeta_x:= x+\delta_x^{-1}\nabla g(x).
\end{aligned}
\end{equation*}

Therefore,
\begin{equation*}\label{eq:skew}
\log\frac{\mathcal N(y\mid0,\Sigma_0)\,q(x\mid y)}{\mathcal N(x\mid0,\Sigma_0)\,q(y\mid x)}
\;=\; \tilde h(x,y)-\tilde h(y,x),
\end{equation*}
with
\begin{equation*}\label{eq:h-tilde}
\begin{aligned}
\tilde h(x,y)
&=
\frac12\frac{\delta_x}{\kappa}\Bigl\|y-x-\frac{\gamma}{\delta_x}\nabla g(x)\Bigr\|^2
-\frac12\sum_{i=1}^d\log(\lambda_i\delta_x+1)
-\frac12\frac{\gamma}{2}\,\bar\zeta_x^\top(\Lambda+\delta_x^{-1}I)^{-1}\bar\zeta_x
+\frac{\gamma}{4}\,x^\top\Sigma_0^{-1}x.
\end{aligned}
\end{equation*}

\subsection{Proof of Theorem 1 
}
\begin{proof}
    First, by the assumption on $\mu(x)$, $P(y|x)$ is $\phi$-irreducible and aperiodic. We check the Lyapunov inequality
    $$
     \int P(dy|x)V(y)\leq \eta V(x)+b\mathbb{I}_C(x)
    $$
    where $\eta<1$ , $b<\infty$ and $C$ is a small set. 
    First,
    \begin{align}\label{eq:pv-v}
      \frac{PV(x)}{V(x)} &= \int P(dy|x)V(y)/V(x) \nonumber\\ &= \int \frac{V(y)}{V(x)}q(y|x)\alpha(x,y)dy+\Big[1-\int q(z|x)\alpha(x,z)dz\Big]\nonumber\\
      &\leq \int\frac{V(y)}{V(x)}q(y|x)dy + \int_{R_R(x)}q(y|x)(1-\alpha(x,y))dy .
    \end{align}
   Let $\lambda_i(A)$ be the $i$-th eigenvalue of matrix $A$. Then, the first term of \eqref{eq:pv-v} becomes
   \begin{align}\label{eq:bound-first-term}
    &\int\frac{V(y)}{V(x)}q(y|x)dy\nonumber\\ &= \frac{1}{(2\pi)^{-d/2}|\Sigma_x|^{-1/2}}\int \exp\{k(y^\top y-x^\top x)\}\exp\{-\frac{1}{2}(y-\mu(x))^\top (\Sigma_x)^{-1}(y-\mu(x))\}dy \nonumber \\
    & = \frac{1}{|I-2k\Sigma_x|^{-1/2}}\exp\{k \mu(x)^\top[I-2k\Sigma_x]^{-1} \mu(x)-kx^\top x\}\nonumber \\
    & \leq \lambda_{\min}(I-2k\Sigma_x)^{-d/2}\exp\{-k(\Vert x\Vert^2 -\Vert \mu(x)\Vert^2 \cdot \Vert[I-2k\Sigma_x]^{-1}\Vert)\}\nonumber\\
    & \leq (1-2k \lambda_{\max}(\Sigma_x))^{-d/2}\exp\{-k(\Vert x\Vert^2 -\frac{\Vert \mu(x)\Vert^2}{1-2k\Vert\Sigma_x\Vert}\} \nonumber\\
    & \leq (1-s)^{-d/2}\exp\{-k(\Vert x\Vert^2 -\frac{\Vert \mu(x)\Vert^2}{1-s})\}.
\end{align}
Note that the penultimate inequality holds because of the Neumann series and that in the last inequality we used $\lambda_{\max}(\Sigma_x) = \Vert \Sigma_x\Vert$. From (A1), when $\Vert x\Vert\rightarrow \infty$, \eqref{eq:bound-first-term} is less than $1$. Also, from (A2
), the second term of \eqref{eq:pv-v} converges to $0$. Therefore we have
$$
    \lim_{\Vert x\Vert\rightarrow\infty} \frac{PV(x)}{V(x)} < 1.
$$
Since the compact sets are small, $P$ is $V$-uniformly geometrically ergodic and from Theorem 16.0.2 of \cite{Meyn2009MarkovEdition},
$$
\Vert P(y|x)^N-\pi(x)\Vert_V \leq C\rho^N V(x)
$$
holds. This guarantees convergence to the target $\pi$ at an exponential rate $\rho \in (0,1)$, which reflects the transition kernel's efficiency in exploring the target space. The error bound depends globally on the constant $C$ and locally on the initial drift penalty $V(x)$.
\end{proof}

\subsection{Proof of Theorem 
2 
}
\begin{proof} Our proof adapts part of the argument from the proof of Theorem 15.4.1 in \cite{Meyn2009MarkovEdition}.
From the Lyapunov inequality, for some constant $b$, we have
$$
    PV(x)\leq V(x)+b\leq(1+b)V(x)
$$
and then
\begin{align*}
    \Delta_{N} &= |P \hat{F}_N - \hat{F}_N - (-F + \pi(F))| \nonumber \\
    & = \Big|PF-\pi(F) + \sum_{n=2}^{N+1} (P^n F-\pi(F)) - [F-\pi(F) + \sum_{n=1}^{N} (P^n F-\pi(F))] - [-F + \pi(F)]\Big| \nonumber \\
    & = |P^{N+1}F - \pi(F)| \nonumber\\
    & = |P^NPF-\pi(PF)|\nonumber\\
    &\leq \Vert P^N-\pi \Vert_{(1+b)V}\nonumber\\
    & = (1+b)\Vert P^N-\pi \Vert_V\nonumber\\
    & \leq (1+b)C\rho^N V(x)
\end{align*}
\end{proof}
\subsection{Proof of Proposition 3 
}
\begin{proof}
    From the proof of Theorem 2 
    and by using Taylor expansion, we have
    \begin{align*}
    \Delta_{N-1} = | P^N F - \pi(F) | & = \Big |\int [\mathcal{N}(y|\beta^N x+(1-\beta^N)\mu, (1-\beta^{2N})\Sigma) - \mathcal{N}(y|\mu,\Sigma)]F(y)dy \Big |\nonumber\\
    & = \Big |\int [\beta^N (y-\mu)^\top \Sigma^{-1}(x-\mu) + O(\beta^{2N})]\mathcal{N}(y|\mu,\Sigma)F(y)dy \Big |\nonumber\\
    & \leq |\beta|^{N} \int |F(y) (y-\mu)^\top \Sigma^{-1}(x-\mu)|\mathcal{N}(y|\mu,\Sigma)dy + O(\beta^{2N})\nonumber\\
    & \leq |\beta|^{N} \Vert x-\mu \Vert\int |F(y)| \cdot\Vert(y-\mu)^\top \Sigma^{-1}\Vert \mathcal{N}(y|\mu,\Sigma)dy + O(\beta^{2N})\nonumber\\
    & = B|\beta|^{N-1} \Vert x-\mu \Vert+ O(\beta^{2N})
\end{align*}
where $B = |\beta|\int |F(y)|\cdot \Vert(y-\mu)^\top \Sigma^{-1}\Vert \mathcal{N}(y|\mu,\Sigma)dy$.
\end{proof}

\subsection{Proof of Theorem 3 \label{app:thr3} }
\begin{proof}
    The proof is directly following the proof of Theorem 2 and Therorem 3 of \cite{dellaportas2012control}. Consider a bivariate chain $Z_n = (X_n,Y_n)$. The marginal process $\{Y_n\}$ can be regarded as a Markov chain consisting of the accepted samples, that is, it is driven by the same proposals as $\{X_n\}$ but having a Gaussian invariant target. From Theorem 1 and the proof of Theorem 3 of \cite{dellaportas2012control}, the chain is ergodic. Since $F$ satisfies (A3), and the control variate involves a finite summation, the LLN and CLT hold according to the proof of the Theorem 2 of \cite{dellaportas2012control},  and from the proof of the Theorem 3 of \cite{dellaportas2012control}, the coefficients $\hat{\theta}_1$ and $\hat{\theta}_2$ are consistent.
\end{proof}

\subsection{Proof of Theorem 4 
}
\begin{proof}
    Recalling the definition of the speed measure given in \cite{titsias2019gradient}
    $$
        s = \int \pi_d(x)\exp\{\kappa \mathcal{H}_{q(y|x)}\}\Big(\int \alpha(x,y)q(y|x)dy\Big)dx,
    $$
    we first simplify the integral and then implement the convergence of the acceptance rate. The exponential of the entropy is
    $$
        \exp\{\kappa \mathcal{H}_{q(y|x)}\} = \exp\{-\kappa \int q(y|x)\log q(y|x)dy\} = (2 \pi e(2\gamma - \gamma^2)E_{f}[-\nabla^2 g(x)]^{-1})^{\kappa d/2}
    $$
    which is independent of position $x$. Therefore, the speed measure becomes
    \begin{align*}
        s &= \int \pi_d(x)\exp\{\kappa \mathcal{H}_{q(y|x)}\}\Big(\int \alpha(x,y)q(y|x)dy\Big)dx \\
        &= C (2\gamma - \gamma^2)^{\kappa d/2}E_{\pi_d,q}[\alpha(x,y)]
    \end{align*}
    where $C =(2 \pi e E_{f}[-\nabla^2 g(x)]^{-1})^{\kappa d/2}$ is a constant. From Lemma 1 and Lemma 3 of \cite{roberts1998optimal}, there exists a sequence of sets $H_d \in R^d$ such that
    $$
        \lim_{d \rightarrow \infty} \sup_{x^d\in H_d}\Big\vert E_{q}[\alpha(x^d,y)]-2 \Phi(-\frac{\epsilon K \sqrt{d}\gamma^{3/2}}{2})  \Big\vert =0
    $$
    where     $$K = \sqrt{\frac{5}{12}E_{f}[h'''(x)^2]+\frac{1}{4}Var_f[h''(x)]}>0$$
    is derived from the Taylor expansion with respect to both $\gamma$ and $\epsilon$ around $0$ via MATHEMATICA 13.3 software. Thus, for any constant $M >0$, there exists a sufficient large $d$, such that 
    $$
        \Big\vert E_{\pi_d, q}[\alpha(x,y)]-2 \Phi(-\frac{\epsilon K \sqrt{d}\gamma^{3/2}}{2})  \Big\vert\leq\sup_{x^d\in H_d}\Big\vert E_{q}[\alpha(x^d,y)]-2 \Phi(-\frac{\epsilon K \sqrt{d}\gamma^{3/2}}{2})  \Big\vert < M
    $$
    Then, it is equivalent to maximize the lower bound of the speed measure
    $$
        s' = (2\gamma-\gamma^2)^{\kappa d/2}\Big(2\Phi(-\frac{\epsilon K \sqrt{d} \gamma^{3/2}}{2})-M\Big)
    $$
    where we can choose a pair of $M$ and $d$ and a sufficiently small $\epsilon$ such that $s'>0$, and the optimization problem becomes
    $$
        \gamma^* = \operatorname*{argmax}_{\gamma} ~ (2\gamma-\gamma^2)^{\kappa d/2}\Big(2\Phi(-\frac{\epsilon K \sqrt{d} \gamma^{3/2}}{2})-M\Big).
    $$
\end{proof}


\end{document}